\documentclass[10pt,twocolumn,letterpaper]{article}

\usepackage{cvpr}              
\usepackage {amsmath}
\usepackage {stfloats}
\usepackage {indentfirst}
\usepackage {multirow}
\usepackage {adjustbox}
\usepackage {makecell}
\usepackage {multicol}
\usepackage {subcaption}
\usepackage {stfloats}
\usepackage[accsupp]{axessibility}  

\usepackage[dvipsnames]{xcolor}

\definecolor{cvprblue}{rgb}{0.21,0.49,0.74}
\usepackage[pagebackref,breaklinks,colorlinks,citecolor=cvprblue]{hyperref}
\allowdisplaybreaks[4]
\def\paperID{12065} 
\def\confName{CVPR}
\def\confYear{2024}

\makeatletter
\def\@fnsymbol#1{\ensuremath{\ifcase#1\or \dagger\or \ddagger\or
   \mathsection\or \mathparagraph\or \|\or **\or \dagger\dagger
   \or \ddagger\ddagger \else\@ctrerr\fi}}
\makeatother
\title{Navigate Beyond Shortcuts: \\ Debiased Learning through the Lens of Neural Collapse}

\author{
Yining Wang, Junjie Sun, Chenyue Wang, Mi Zhang\thanks{Corresponding Author.} , Min Yang\\
School of Computer Science, Fudan University, China\\
{\tt\small \{ynwang22@m.,jjsun22@m.,wangcy23@m.,mi\_zhang@,m\_yang@\}fudan.edu.cn}
}
\begin{document}
\maketitle
\begin{abstract}
Recent studies have noted an intriguing phenomenon termed Neural Collapse, that is, when the neural networks establish the right correlation between feature spaces and the training targets, their last-layer features, together with the classifier weights, will collapse into a stable and symmetric structure. In this paper, we extend the investigation of Neural Collapse to the biased datasets with imbalanced attributes. We observe that models will easily fall into the pitfall of shortcut learning and form a biased, non-collapsed feature space at the early period of training, which is hard to reverse and limits the generalization capability. To tackle the root cause of biased classification, we follow the recent inspiration of prime training, and propose an avoid-shortcut learning framework without additional training complexity. With well-designed shortcut primes based on Neural Collapse structure, the models are encouraged to skip the pursuit of simple shortcuts and naturally capture the intrinsic correlations. Experimental results demonstrate that our method induces better convergence properties during training, and achieves state-of-the-art generalization performance on both synthetic and real-world biased datasets.
\end{abstract}
\section{Introduction}
\label{sec:1}
\begin{figure}[htbp]
    \centering
\includegraphics[width=\columnwidth]{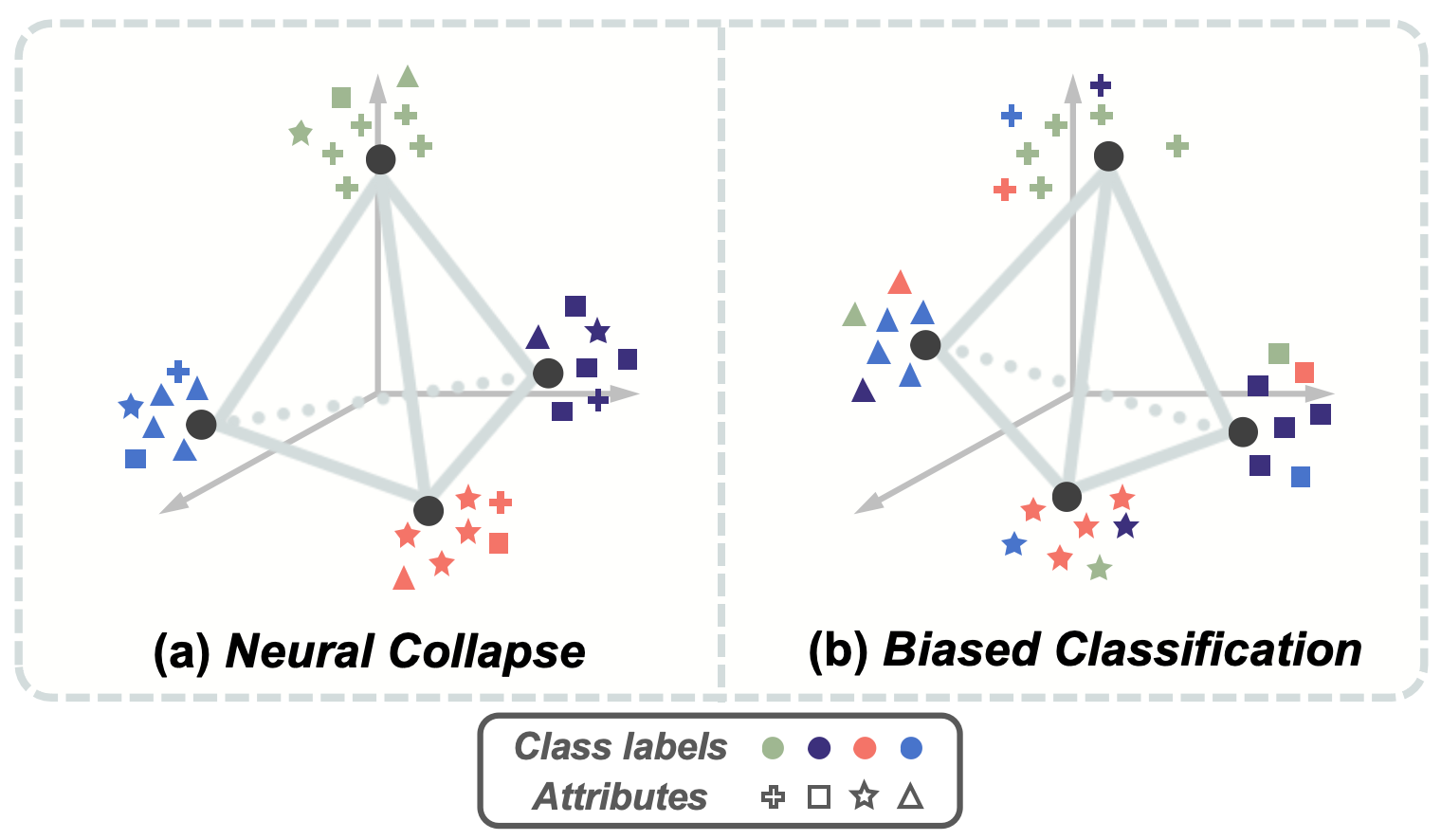}
    \caption{Illustration of (a) \textbf{Neural Collapse} phenomenon on balanced datasets, where the simplex ETF structure maximizes the class-wise angles, and (b) \textbf{Biased classification} on datasets with imbalanced attributes, where the model takes the shortcut of attributes to make predictions and fails to collapse into the simplex ETF. The color of points represents different class labels and the shape of points represents different attributes.}
    \label{fig:nc}
\vspace{-1.0em}
\end{figure}

When the input-output correlation learned by a neural network is consistent with its training target, the last-layer features and classifier weights will attract and reinforce each other, forming a stable, symmetric and robust structure. Just as the Neural Collapse phenomenon discovered by Papyan et al. \cite{papyan2020prevalence}, at the terminal phase of training on balanced datasets, a model will witness its last-layer features of the same class converge towards the class centers, and the classifier weights align to these class centers correspondingly. The convergence will ultimately lead to the collapse of feature space into a simplex equiangular tight frame (ETF) structure, as illustrated in Fig. \ref{fig:nc}(a). The elegant structure has demonstrated its efficacy in enhancing the generalization, robustness, and interpretability of the trained models \cite{papyan2020prevalence, fang2021exploring}. Therefore, a wave of empirical and theoretical analysis of Neural Collapse has been proposed \cite{dang2023neural, han2021neural, galanti2022role, peifeng2023feature, rangamani2023feature, yang2023neurons, yaras2022neural}, and a series of studies have adopted the simplex ETF as the optimal geometric structure of the classifier, to guide the maximized class-wise separation in class-imbalanced training \cite{li2023no, zhong2023understanding, yang2022neural, xie2023neural, yang2022inducing}. 

However, in practical visual recognition tasks, besides the challenge of inter-class imbalance, we also encounter intra-class imbalance, where the majority of samples are dominated by the bias attributes (e.g., some misleading contents such as background, color, texture, etc.). For example, the widely used LFW dataset \cite{huang2008labeled} for facial recognition has been demonstrated severely imbalanced in gender, age and ethnicity \cite{deviyani2022assessing}. A biased dataset often contains a majority of \textit{bias-aligned} samples and a minority of \textit{bias-conflicting} ones. The prevalent bias-aligned samples exhibit a strong correlation between the ground-truth labels and bias attributes, while the scarce bias-conflicting samples have no such correlation. Once a model relies on the simple but spurious \textit{shortcut} of bias attributes for prediction, it will ignore the intrinsic relations and struggle to generalize on out-of-distribution test samples. The potential impact of biased classification may range from political and economic disparities to social inequalities within AI systems, 
as emphasized in EDRi's latest report \cite{balayn2021beyond}.

Therefore, the fundamental solution to biased classification lies in deferring, or ideally, preventing the learning of shortcut correlations. However, previous \textit{debiased learning} methods rely heavily on additional training expenses. For example, a bias-amplified auxiliary model is often adopted to identify and up-weight the bias-conflicting samples \cite{nam2020learning, lee2023revisiting, kim2022learning}, or employed to guide the input-level and feature-level augmentations \cite{hwang2022selecmix, lee2021learning, lim2023biasadv}. Some disentangle-based debiasing methods, from the perspective of causal intervention \cite{wang2021causal, zhang2023benign} or Information Bottleneck theory \cite{tartaglione2021end}, also require large amounts of contrastive samples or pre-training process to disentangle the biased features, significantly increasing the burden of debiased learning.

In this paper, we extend the investigation of Neural Collapse to the biased visual datasets with imbalanced attributes. Through the lens of Neural Collapse, we observe that models prioritize the period of \textit{shortcut learning}, and quickly form the biased feature space based on misleading attributes at the early stage of training. After the bias-aligned samples reach zero training error, the intrinsic correlation within bias-conflicting samples will then be discovered. However, due to i) the scarcity of bias-conflicting samples and ii) the stability of the established feature space, the learned shortcut correlation is challenging to reverse and eliminate. The mismatch between bias feature space and the training target induces inferior generalizability, and hinders the convergence of Neural Collapse, as shown in Fig. \ref{fig:nc}(b).

To achieve efficient model debiasing, we follow the inspiration of \textit{prime training}, and encourage the model to skip the active learning of shortcut correlations. The \textit{primes} are often provided as additional supervisory signals to redirect the model's reliance on shortcuts, which helps improve generalization in image classification and CARLA autonomous driving \cite{wen2022fighting}. To rectify models’ attention on the intrinsic correlations, we define the primes with a training-free simplex ETF structure, which approximates the ``optimal" shortcut features and guides the model to pursue unbiased classification from the beginning of training. Our method is free of auxiliary models or additional optimization of prime features. Experimental results also substantiate its state-of-the-art debiasing performance on both synthetic and real-world biased datasets. 

Our contributions are summarized as follows:
\begin{itemize}
    \item For the first time, we investigate the Neural Collapse phenomenon on biased datasets with imbalanced attributes. Through the empirical results of feature convergence, we analyze the shortcut learning stage of training, as well as the fundamental issues of biased classification.
    \item We propose an efficient avoid-shortcut training paradigm, which introduces the simplex ETF structure as prime features, to rectify models’ attention on the intrinsic correlations.
    \item We demonstrate the state-of-the-art debiasing performance of our method on 2 synthetic and 3 real-world biased datasets, as well as the better convergence properties of debiased models. 
\end{itemize}

\section{Related Works}
\label{sec:2}
\noindent \textbf{Debiased Learning.}
\label{sec:2.1}
Extensive efforts have been dedicated to model debiasing, but they are significantly limited by additional training costs. Recent advances can be divided into three categories: reweight-based, augmentation-based, and disentangle-based. Based on the easy-to-learn heuristic of biased features, reweight-based approaches require pre-trained bias-amplified models to identify and emphasize the bias-conflicting training samples \cite{nam2020learning, lee2023revisiting, kim2022learning}. Augmentation-based approaches, with the guidance of explicit bias annotations, conduct image-level and feature-level augmentations to enhance the diversity of training datasets \cite{hwang2022selecmix, lee2021learning, lim2023biasadv}. Other disentangle-based approaches attempt to remove the bias-related part of features, from the perspective of Information Bottleneck theory \cite{tartaglione2021end} or causal intervention \cite{wang2021causal, zhang2023benign}, but at the cost of substantial contrastive samples. Additionally, model debiasing is also well studied in graph neural networks \cite{fan2022debiasing, zhang2023debiasing}, language models \cite{guo2022auto, lyu2023feature} and multi-modal tasks \cite{wen2021debiased, hirota2023model}.

\noindent \textbf{Neural Collapse.} 
Discovered by Papyan et al. \cite{papyan2020prevalence}, the Neural Collapse phenomenon reveals the convergence of the last-layer feature space to an elegant geometry.
At the terminal phase of training on balanced datasets, the feature centers and classifier weights will collapse together into the structure of a simplex ETF, which is illustrated in Section \ref{sec:3.1}. Recent works have dug deeper into the phenomenon and provided theoretical supports under different constraints or regularizations \cite{dang2023neural, yaras2022neural, han2021neural}, as well as empirical studies of intermediate features and transfer learning \cite{rangamani2023feature, yang2023neurons, peifeng2023feature, galanti2022role}. Considering the class-imbalanced datasets, Fang et al. \cite{fang2021exploring} point out the Minority Collapse phenomenon, where features of long-tailed classes will merge together and be hard to classify. As a remedy, they fix the classifier as an ETF structure during training, which guarantees the optimal geometric property in imbalanced learning \cite{yang2022inducing}, semantic segmentation \cite{zhong2023understanding}, and federated learning \cite{yang2022neural}. To take a step further, our work fills the gap of Neural Collapse analysis on biased datasets with shortcut correlations. 

\noindent \textbf{Avoid-shortcut Learning.}
The recent inspiration of avoid-shortcut learning aims to postpone, or even prevent the learning of shortcut relations in model training. With well-crafted contrastive samples \cite{robinson2021can, shinoda2023shortcut, saranrittichai2022overcoming} or artificial shortcut signals \cite{wen2022fighting, zhang2023benign}, avoid-shortcut learning has demonstrated its efficacy in image classification, autonomous driving and question answering models. One of the representative methods is named prime training, which provides richer supervisory signals of key input features (i.e., \textit{primes}) to guide the establishment of correct correlations, therefore improving generalization on OOD samples \cite{wen2022fighting}. In this work, we leverage the approximated ``optimal" shortcuts as primes to encourage the models to bypass shortcut learning.
 
\section{Preliminaries}
\subsection{Neural Collapse Phenomenon}
\label{sec:3.1}
Consider a biased dataset $\mathcal{D}$ with $K$ classes of training samples, we denote $\mathbf{x}_{k,i}$ as the $i$-th sample of the $k$-th class and $\mathbf{z}_{k,i} \in \mathbb{R}^d$ as its corresponding last-layer feature. A linear classifier with weights $\mathbf{W} = [\mathbf{w}_1, ..., \mathbf{w}_K] \in \mathbb{R}^{d \times K}$ is trained upon the last-layer features to make predictions.

The Neural Collapse (NC) phenomenon discovered that, when neural networks are trained on balanced datasets, the correctly learned correlations will naturally lead to the convergence of feature spaces. Given enough training steps after the zero classification error, the last-layer features and classifier weights will collapse to the vertices of a simplex equiangular tight frame (ETF), which is defined as below.

\noindent \textbf{Definition 1 (Simplex Equiangular Tight Frame)} A collection of vectors $\mathbf{m}_k \in \mathbb{R}^d$, $k=1,2,...,K, d \ge K-1$ is said to be a $k$-simplex equiangular tight frame if:
\begin{equation}
\label{eq1}
    \mathbf{M}=\sqrt{\frac{K}{K-1}} \mathbf{P} (\mathbf{I}_K - \frac{1}{K} \mathbf{1}_K \mathbf{1}_K^\mathrm{T})
\end{equation}
where $\mathbf{M}=[\mathbf{m}_1, ..., \mathbf{m}_K] \in \mathbb{R}^{d \times K}$, and $\mathbf{P} \in \mathbb{R}^{d \times K}$ is an orthogonal matrix which satisfies $\mathbf{P}^\mathrm{T} \mathbf{P} = \mathbf{I}_K$, with $\mathbf{I}_K$ denotes the identity matrix and $\mathbf{1}_K$ denotes the all-ones vector. Within the ETF structure, all vectors have the maximal pair-wise angle of $-\frac{1}{K-1}$,  namely the maximal equiangular separation.

Besides the convergence to simplex ETF structure, the Neural Collapse phenomenon could be concluded as the following properties during the terminal phase of training:

\noindent\textbf{NC1: Variability collapse.} The last-layer features $\mathbf{z}_{k,i}$ of the same class $k$ will collapse to their class means $\mathbf{\overline{z}}_k = {\rm Avg}_i \{\mathbf{z}_{k,i}\}$, and the within-class variation of the last-layer features will approach 0.

\noindent\textbf{NC2: Convergence to simplex ETF.} The normalized class means will collapse to the vertices of a simplex ETF. We denote the global mean of all last-layer features as $\mathbf{z}_G={\rm Avg}_{i,k} \{\mathbf{z}_{k,i}\}, k \in [1,...,K]$ and the normalized class means as $\mathbf{{\tilde{z}}}_k = (\mathbf{z}_k - \mathbf{z}_G)/ \lVert\mathbf{z}_k - \mathbf{z}_G \rVert$, which satisfies Eq.\ref{eq1}.

\noindent\textbf{NC3: Self duality.} The classifier weights $\mathbf{w}_k$ will align with the corresponding normalized class means $\mathbf{{\tilde{z}}}_k$, which satisfies $\mathbf{{\tilde{z}}}_k = \mathbf{w}_k / ||\mathbf{w}_k||$.

\noindent\textbf{NC4: Simplification to nearest class center.} After convergence, the model's prediction will collapse to simply choosing the nearest class mean to the input feature (in standard Euclidean distance). The prediction of $\mathbf{z}$ could be denoted as $\arg\max_k \langle \mathbf{z}, \mathbf{w}_k \rangle = \arg\min_k ||\mathbf{z} - \mathbf{\overline{z}}_k||$.

\begin{figure*}[htbp]
    \centering
    \includegraphics[width=\textwidth]{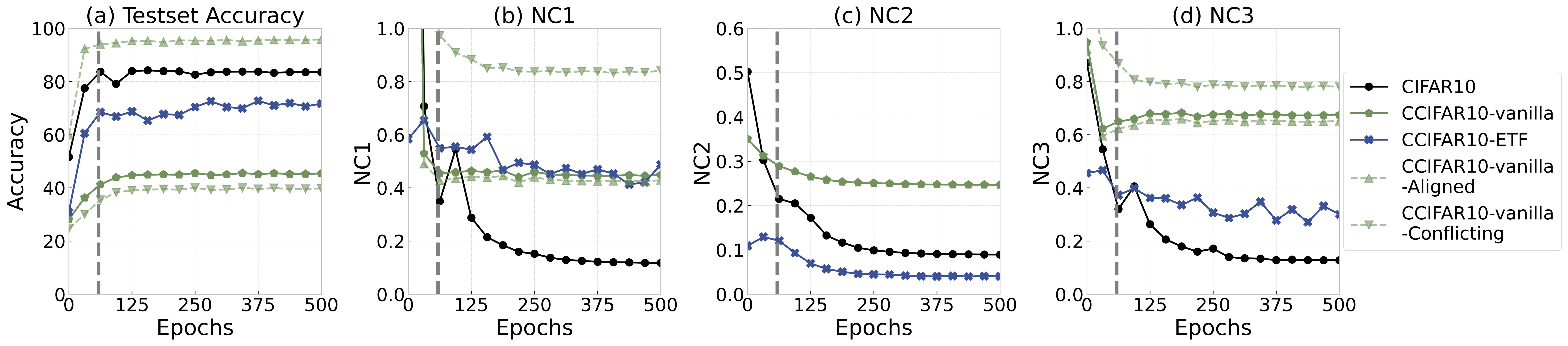}
    \caption{Comparison of \textbf{(a) testset accuracy} and \textbf{(b-d) Neural Collapse metrics} on unbiased (CIFAR-10) and synthetic biased (Corrupted CIFAR-10 with the bias ratio of 5.0\%) datasets. All vanilla models are trained with standard cross-entropy loss for 500 epochs. The postfix \textit{-Aligned} and \textit{-Conflicting} indicate the results of bias-aligned and bias-conflicting samples respectively. The NC1 metric evaluates the convergence of same-class features, NC2 evaluates the difference between the feature space and a simplex ETF, and NC3 measures the duality between feature centers and classifier weights. The vertical dashed line at the epoch of 60 divides two stages of training. }
    \label{fig:NC}
\end{figure*}
\subsection{Neural Collapse Observation on Biased Dataset}
\label{sec:3.2}
Besides the findings on balanced datasets, some studies have explored Neural Collapse under the class-imbalanced situation \cite{yang2022inducing, fang2021exploring}. Taking a step further, we investigate the phenomenon on biased datasets with imbalanced attributes, to advance the understanding of biased classification. To examine the convergence of last-layer features and classifier weights, we compare the metrics of Neural Collapse on both unbiased and synthetic biased datasets. As shown in Fig. \ref{fig:NC}, we report the result of NC1-NC3, which corresponds to the first three convergence properties in Section \ref{sec:3.1} and respectively evaluates the convergence of same-class features, the structure of feature space and self-duality. The details of NC metrics are concluded in Tab. \ref{tab:metrics}.

\begin{table*}
    \centering
    \begin{adjustbox}{width=0.85\textwidth}
    \begin{tabular}{cl}
    \toprule[1.5pt]
    \makebox[0.1\textwidth][c]{Metrics} & 
    \makebox[0.6\textwidth][c]{Computational details} \\
    \hline
        NC1 & ${\rm \mathcal{NC}_1}=\frac{1}{K} {\rm Tr}(\Sigma_W\Sigma_B^\dag)$, where ${\rm Tr}$ is the trace of matrix and $\Sigma_B^\dag$ denotes the pseudo-inverse of $\Sigma_B$ \\
        & $\Sigma_B = \frac{1}{K} \sum\limits_{k \in [K]} (\mathbf{\overline{z}}_k - \mathbf{z}_G)(\mathbf{\overline{z}}_k - \mathbf{z}_G)^\mathrm{T}$,  $\Sigma_W=\frac{1}{K}\sum\limits_{k \in [K]} \frac{1}{n_k} \sum\limits_{i=1}^{n_k} (\mathbf{z}_{k,i} - \mathbf{\overline{z}}_k)(\mathbf{z}_{k,i} - \mathbf{\overline{z}}_k)^\mathrm{T}$ \\
    \hline
        NC2 & ${\rm \mathcal{NC}_2}=  \Big\lVert \frac{\mathbf{W}\mathbf{W}^\mathrm{T}}{||\mathbf{W}\mathbf{W}^\mathrm{T}||_\mathcal{F}} - \frac{1}{\sqrt{K-1}}(\mathbf{I}_K-\frac{1}{K}\mathbf{1}_K \mathbf{1}_K^\mathrm{T}) \Big\rVert_\mathcal{F}$\\
    \hline
        NC3 & ${\rm \mathcal{NC}_3}=  \Big\lVert \frac{\mathbf{W}\mathbf{\overline{Z}}}{||\mathbf{W}\mathbf{\overline{Z}}||_\mathcal{F}} - \frac{1}{\sqrt{K-1}}(\mathbf{I}_K-\frac{1}{K}\mathbf{1}_K \mathbf{1}_K^\mathrm{T}) \Big\rVert_\mathcal{F}$, where $\mathbf{\overline{Z}}=[\mathbf{\overline{z}}_1 - \mathbf{z}_G, ..., \mathbf{\overline{z}}_K - \mathbf{z}_G]$ \\
    \bottomrule[1.5pt]
    \end{tabular}
    \end{adjustbox}
    \caption{The metrics of evaluating the Neural Collapse phenomenon, which are generally adopted in previous studies \cite{zhu2021geometric, yang2023neurons, papyan2020prevalence}. $\Vert \cdot \Vert_{\mathcal{F}}$ denotes the Frobenius norm, $\mathbf{I}_K$ is the identity matrix and $\mathbf{1}_K$ is the all-ones vector.}
    \label{tab:metrics}
\vspace{-1.0em}
\end{table*}

When trained on unbiased datasets ({\textbf{black lines}} in Fig. \ref{fig:NC}), the model displays the expected convergence properties, with metrics NC1-NC3 all converge to zero. We owe the elegant collapse phenomenon to the right correlation between the feature space and training objective, which is also supported by the analysis of benign global landscapes \cite{zhu2021geometric, zhou2022all}. 

However, when trained on biased datasets, the training process exhibits two stages: first the \textit{shortcut learning} period and then the \textit{intrinsic learning} period, as divided by the vertical dashed line. During the shortcut learning period, the accuracy of bias-aligned samples increases quickly, and the NC1-NC3 metrics show a rapid decline (\textcolor[RGB]{113,147,91}{\textbf{green lines with $\blacktriangle$}} in Fig. \ref{fig:NC}). It indicates that when simple shortcuts exist in the training distribution, the model will quickly establish its feature space based on the bias attributes, and exhibit a converging trend towards the simplex ETF structure. 

After the bias-aligned samples approach zero error, the model turns to the period of \textit{intrinsic learning}, which focuses on the intrinsic correlations within bias-conflicting samples to further reduce the empirical loss. However, although their final loss reduces to zero, the bias-conflicting samples still display low accuracy and poor convergence results (\textcolor[RGB]{113,147,91}{\textbf{green lines with $\blacktriangledown$}}). It implies that the intrinsic learning period merely induces the over-fitting of bias-conflicting samples and does not benefit in generalization. We attribute the failure of collapse to the early establishment of shortcut correlations. Once the biased feature space is established based on misleading attributes, rectifying it becomes challenging, particularly with scarce bias-conflicting samples. In the subsequent training steps, the misled features of bias-conflicting samples will hinder the convergence of same-class features, thereby halting the converging trend towards the simplex ETF structure and leading to a non-collapsed, sub-optimal feature space.  

To break the curse of shortcut learning, we turn the tricky shortcut into a training prime, which effectively guides the models to focus on intrinsic correlations and form a naturally collapsed feature space (\textcolor[RGB]{59,82,152}{\textbf{blue lines}} in Fig. \ref{fig:NC}). Our method is presented in the following sections.

\section{Methodology}
\begin{figure*}[htbp]
    \centering
\includegraphics[width=0.9\textwidth]{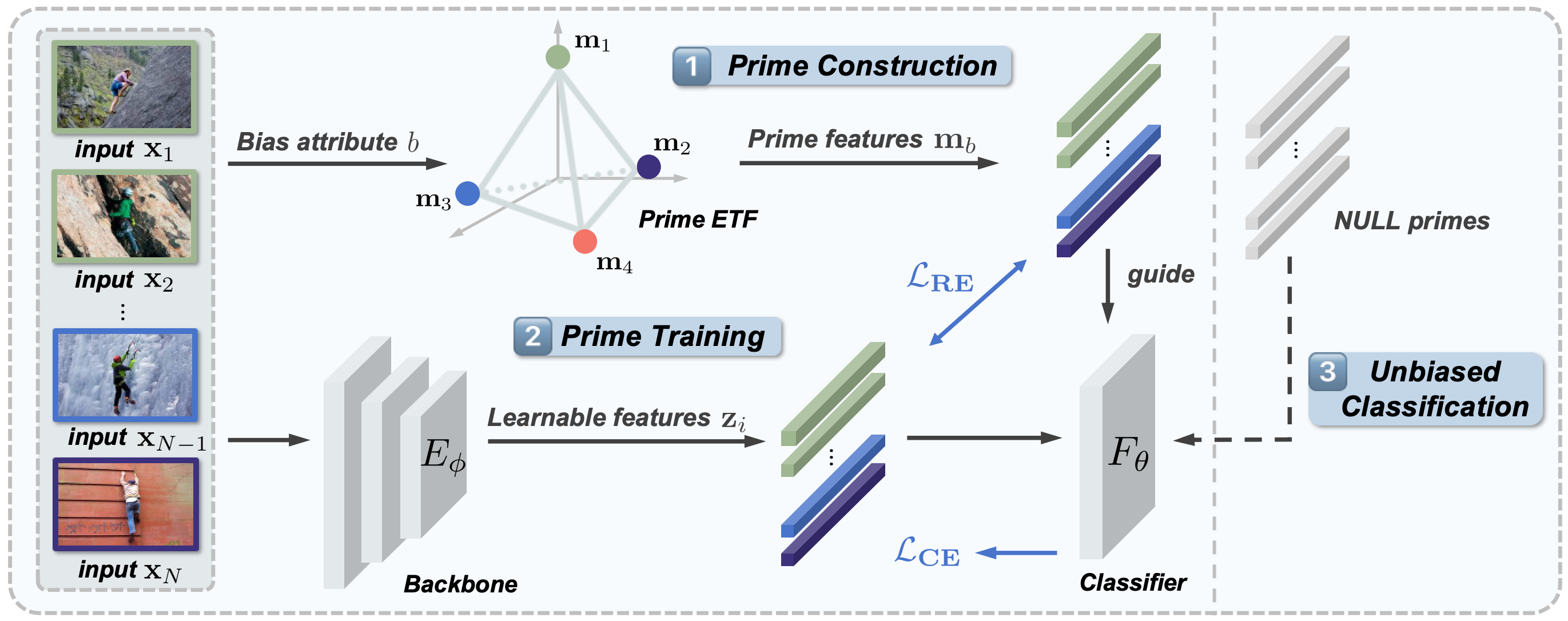}
\caption{The illustration of our method. We take the class \textit{climbing} from BAR \cite{nam2020learning} as an example, which contains samples of human climbing but with the bias attribute of different backgrounds (as indicated with the color of image frames). The framework contains: 1) \textbf{Prime Construction}: Before training, a randomly initialized ETF structure is constructed as the shortcut primes, and 2) During  \textbf{Prime Training}, the prime features $\mathbf{m}_b$ are retrieved based on the bias attribute $b$ of the input samples, to guide the optimization of learnable features $\mathbf{z}$ towards the intrinsic correlations. The classifier $F_\theta$ will take both the learnable features and fixed prime features to make predictions. 3) In \textbf{Unbiased Classification}, the prime features are assigned as null vectors to evaluate the debiased model on test distributions.}
\label{fig:framework}
\end{figure*}
\subsection{Motivation}
Following the previous analysis, we highlight the importance of redirecting the model’s emphasis from simple shortcuts to intrinsic relations. Since the models can be easily misled by shortcuts in the training distribution, is it feasible to supply a ``perfectly learned" shortcut feature, to deceive the models into skipping the active learning of shortcuts, and directly focusing on the intrinsic correlations?

We observe in Fig. \ref{fig:NC} that the NC1-NC3 metrics show a rapid decrease during shortcut learning, but remain stable in the subsequent training epochs. However, if the training distribution does follow the shortcut correlation (with no obstacle from bias-conflicting samples), the convergence will end up with the optimal structure of simplex ETF, just as the results on unbiased datasets. This inspires us to approximate the ``perfectly learned" shortcut features with a simplex ETF structure, which requires no additional training and represents the optimal geometry of feature space.

Therefore, following the outstanding performance of prime training in OOD generalization \cite{wen2022fighting}, we introduce the approximated ``perfect" shortcuts as the primes for debiased learning. The provided shortcut primes are constructed with a training-free simplex ETF structure, which encourages the models to directly capture the intrinsic correlations, therefore exhibit superior generalizability and convergence properties in our experiments.

\subsection{Avoid-shortcut Learning with Neural Collapse}
Building upon our motivation of avoid-shortcut learning, the illustration of the proposed \textbf{ETF-Debias} is shown in Fig. \ref{fig:framework}. The debiased learning framework can be divided into three stages: \textit{prime construction}, \textit{prime training}, and \textit{unbiased classification}. Firstly, a prime ETF will be constructed to approximate the ``perfect" shortcut features. Then during the prime training, the model will be guided to directly capture the intrinsic correlations with the prime training and the prime reinforcement regularization. In evaluation, we rely on the intrinsic correlations to perform unbiased classification. The details are as follows.

\noindent \textbf{Prime construction.} When constructing the prime ETF, we first randomly initialize a simplex ETF as $\mathbf{M} \in \mathbb{R}^{d \times B}$, which satisfies the definition in Eq. \ref{eq1}. 
The dimension $d$ is the same as the learnable features, and the number of vectors in $\mathbf{M}$ is determined by the categories of bias attributes $\mathbf{b}_i \in \{1,...,B\}$, which are pre-defined in the training distribution. 
After initialization, the vertices of prime ETF $[\mathbf{m}_1,...,\mathbf{m}_B]$ are considered as the approximation of the ``perfect" shortcut features for each attribute, which serve as the prime features for avoid-shortcut training. During training, the prime features will be retrieved based on the bias attribute $b$ of each input sample.

\noindent \textbf{Prime training.} During the prime training, we take the end-to-end model architecture with a backbone $E_\phi$ and a classifier $F_\theta$. For the $i$-th input $\mathbf{x}_{i,b}$ with the bias attribute of $b$, we first extract its learnable feature $\mathbf{z}_{i,b}=E_\phi(\mathbf{x}_{i,b})$ with the backbone model, and retrieve its prime feature $\mathbf{m}_b$ based on the bias attribute $b$. The classifier $F_\theta$ will take both the learnable feature $\mathbf{z}_{i,b}$ and the prime feature $\mathbf{m}_b$ to make softmaxed predictions $\hat{\mathbf{y}}=F_\theta(\mathbf{z}_{i,b},\mathbf{m}_b)$. The standard classification objective is defined as:
\begin{align}
    \min_{\phi,\theta} \mathcal{L}_{\rm CE}(\mathbf{x}, \mathbf{y}) = \sum
    _{i=1}^N\mathcal{L}(F_\theta(\mathbf{z}_{i,b},\mathbf{m}_b), \mathbf{y}_{i,b})
\label{loss}
\end{align}
where $\mathbf{y}$ is the ground-truth label. In our implementation, we use the standard cross-entropy loss as $\mathcal{L}$, and concatenate the prime features after the learnable features to perform predictions. 

In essence, we provide a pre-defined prime feature for each training sample based on its bias attribute. The prime features, with a strong correlation with the bias attributes, can be viewed as the optimal solution to shortcut learning. By leveraging the already ``perfect" representation of shortcut correlations, the model will be forced to explore the intrinsic correlations within the training distribution. The prime-guided mechanism targets at the fundamental issue of biased classification, without inducing extra training costs.

\noindent \textbf{Prime reinforcement regularization.} Given the prime features, the model is encouraged to grasp the intrinsic correlation of the training distributions. However, we raise another potential risk that, despite the provided ``perfectly learned" shortcut features, the model may still pursue the easy-to-follow shortcuts, leading to the redundancy between the learnable feature $\mathbf{z}_{i,b}$ and the fixed $\mathbf{m}_b$. We point out that the model may not establish a strong correlation between the prime features and the bias attributes, and continues to optimize the learnable features for the missing connections. 

Therefore, we introduce a \textit{prime reinforcement regularization} mechanism to enhance the model's dependency on prime features. We encourage the model to classify the bias attributes with only the prime features, and the regularization loss is defined as:
\begin{equation}
\label{eq:regularization}
\resizebox{0.9\hsize}{!}{
    $\mathcal{L}_{\rm RE}(\mathbf{x}, \mathbf{b}) = \sum_{i=1}^N \mathcal{L}(F_\theta(\mathbf{z}_{i,b},\mathbf{m}_b)-F_\theta(\mathbf{z}_{i,b},\mathbf{m}_{\rm null}), \mathbf{b})$}
\end{equation}
where $\mathbf{m}_{\rm null}$ is implemented as all-zero vectors with the same dimension as $\mathbf{m}_b$, and $\mathcal{L}$ is the standard cross-entropy loss. In the ablation studies in Section \ref{sec:5.3}, we observe an improved generalization capability across test distributions, as the result of the strengthened reliance on prime features. Regarding the entire framework, we define the overall training objective as:
\begin{equation}
\label{objective}
    \min_{\phi,\theta} \mathcal{L}_{\rm CE}(\mathbf{x}, \mathbf{y}) + \alpha \mathcal{L}_{\rm RE}(\mathbf{x}, \mathbf{b})
\end{equation}
where $\alpha$ is the hyper-parameter to adjust the regularization.

\noindent \textbf{Unbiased classification.} In evaluation, we rely on the intrinsic correlations to perform unbiased classification. Given a test sample $\mathbf{x}_{i,b}$, we extract its learnable feature $\mathbf{z}_{i,b}$ and set its prime feature as $\mathbf{m}_{\rm null}$ to obtain the final output $\hat{\mathbf{y}}=F_\theta(\mathbf{z}_{i,b},\mathbf{m}_{\rm null})$.

\subsection{Theoretical Justification}
\label{sec:4.3}
Based on the analysis of Neural Collapse from the perspective of gradients \cite{zhong2023understanding, yang2022inducing}, we provide a brief theoretical justification for our method. 

With the priming mechanism, we denote the $i$-th feature of the $k$-th class as $\widetilde{\mathbf{z}}_{k,i}=[\mathbf{z}_{k,i}, \mathbf{m}_{i,b}] \in \mathbb{R}^{2 \times d}$, which represents the concatenation of learnable feature $\mathbf{z}_{k,i}$ and prime feature $\mathbf{m}_{i,b}$ based on its bias attribute $b$. To keep the same form, we also denote the classifier weights as $\widetilde{\mathbf{w}}_k=[\mathbf{w}_{k}, \mathbf{a}_k] \in \mathbb{R}^{2 \times d}$, where $\mathbf{w}_{k}$ represents the weight for intrinsic correlations and $\mathbf{a}_k$ represents the one for shortcut correlations. We observe that, due to the fixed prime features during training, $\mathbf{a}_k$ will quickly collapse to the bias-correlated prime features of class $k$, and can be viewed as constant after just a few steps of training. With the definition, the cross-entropy (CE) loss can be written as:
\begin{equation}
\resizebox{\hsize}{!}{
    $\mathcal{L}_{\rm CE}(\widetilde{\mathbf{z}}_{k,i}, \widetilde{\mathbf{w}}_k) = -\log \left( \frac{\exp([\mathbf{z}_{k,i}, \mathbf{m}_{i,b}]^\mathrm{T}[\mathbf{w}_{k}, \mathbf{a}_k])}{\sum_{k'=1}^K \exp([\mathbf{z}_{k,i}, \mathbf{m}_{i,b}]^\mathrm{T}[\mathbf{w}_{k'}, \mathbf{a}_k'])} \right)$}
\label{eq:ce}
\end{equation}We follow the analysis of previous works and compute the gradients of $\mathcal{L}_{\rm CE}$ \textit{w.r.t} both classifier weights and features. 

\noindent \textbf{Gradient \textit{w.r.t} classifier weights.} We first compute the gradient of
$\mathcal{L}_{\rm CE}$ \textit{w.r.t} classifier weights $\mathbf{\widetilde{W}}=[\widetilde{\mathbf{w}}_1...,\widetilde{\mathbf{w}}_K]$:
\setlength\abovedisplayskip{0pt}
\setlength\belowdisplayskip{2pt}
\begin{align}
\label{eq:wrt_w}
    \frac{\partial \mathcal{L}_{\rm CE}}{\partial \widetilde{\mathbf{w}}_k} 
    &= \sum_{i=1}^{n_k} -(1-p_k(\widetilde{\mathbf{z}}_{k,i}))\widetilde{\mathbf{z}}_{k,i} + \sum_{k' \ne k}^K \sum_{j=1}^{n_{k'}} p_k(\widetilde{\mathbf{z}}_{k',j})\widetilde{\mathbf{z}}_{k',j} \nonumber\\
    &\le \underbrace{\sum_{i=1}^{n_k} - (1 - p_k^{(b)}(\mathbf{m}_{i,b}) - p_k^{(l)}(\mathbf{z}_{k,i}) ) \widetilde{\mathbf{z}}_{k,i}}_{\rm pulling \ part} \nonumber\\
    & + \underbrace{\sum_{k' \ne k}^K \sum_{j=1}^{n_{k'}} (p_k^{(b)}(\mathbf{m}_{j,b'}) + p_k^{(l)}(\mathbf{z}_{k',j}) )\widetilde{\mathbf{z}}_{k',j}}_{\rm forcing \ part}
\end{align}
where $p_k^{(l)}$ and $p_k^{(b)}$ are the predicted probabilities for class labels and bias attributes, calculated with softmax:
\setlength\abovedisplayskip{1pt}
\setlength\belowdisplayskip{1pt}
\begin{align}
    p_k^{(l)}(\mathbf{z}_{k,i}) = \frac{\exp (\mathbf{z}_{k,i}^\mathrm{T}\mathbf{w}_{k})}{\sum_{k'=1}^K \exp ([\mathbf{z}_{k,i}, \mathbf{m}_{i,b}]^\mathrm{T}[\mathbf{w}_{k'}, \mathbf{a}_{k'}])}
\label{eq:pk_1}
\end{align}
\setlength\abovedisplayskip{0pt}
\setlength\belowdisplayskip{2pt}
\begin{equation}
    p_k^{(b)}(\mathbf{m}_{i,b}) = \frac{\exp (\mathbf{m}_{i,b}^\mathrm{T}\mathbf{a}_{k})}{\sum_{k'=1}^K \exp ([\mathbf{z}_{k,i}, \mathbf{m}_{i,b}]^\mathrm{T}[\mathbf{w}_{k'}, \mathbf{a}_{k'}])}
\label{eq:pk_2}
\end{equation}

In Eq. \ref{eq:wrt_w}, the gradient \textit{w.r.t} classifier weights are divided into two parts. The \textit{pulling part} is composed of features from the same class that pulls $\mathbf{w}_k$ towards the direction of the $k$-th feature cluster, while the \textit{forcing part} contains the features of other classes and pushes $\mathbf{w}_k$ away from their clusters. The weight factor of each feature $\widetilde{\mathbf{z}}_{k,i}$ represents its influence on the optimization of $\widetilde{\mathbf{w}}_k$, which implicitly plays the role of re-weighting in our method. 

We assume that class $k$ is strongly correlated with bias attribute $b$. As the weight $\mathbf{a}_k$ is observed to collapse quickly to the bias-correlated prime feature $\mathbf{m}_b$, the probability $p_k^{(b)}\propto\exp(\mathbf{m}_{b}^\mathrm{T}\mathbf{a}_{k})$ of bias-aligned samples (with prime features $\mathbf{m}_{b}$) are much greater than that of bias-conflicting samples (with prime features $\mathbf{m}_{b'}$). Thus, with the weight factors in Eq. \ref{eq:wrt_w}, the pulling and forcing effects of bias-aligned samples will be relatively down-weighted, and the impact of bias-conflicting samples will be up-weighted. The re-weighting mechanism of gradient mitigates the tendency of pulling $\widetilde{\mathbf{w}}_k$ towards the center of bias-aligned samples, which alleviates the misdirection of bias attributes. 

\noindent \textbf{Gradient \textit{w.r.t} features.} Similarly, we compute the gradient of $\mathcal{L}_{\rm CE}$ \textit{w.r.t} the feature $\widetilde{\mathbf{z}}_{k,i}$:
\setlength\abovedisplayskip{1pt}
\setlength\belowdisplayskip{2pt}
\begin{align}
    \frac{\partial \mathcal{L}_{\rm CE}}{\partial \widetilde{\mathbf{z}}_{k,i}} 
    &= -(1- p_k(\widetilde{\mathbf{z}}_{k,i})) \mathbf{w}_k
 + \sum_{k' \ne k}^K p_{k'}(\widetilde{\mathbf{z}}_{k,i})\mathbf{w}_{k'} \nonumber\\[1mm]
    &\le \underbrace{-(1 - p_k^{(b)}(\mathbf{m}_{i,b}) - p_k^{(l)}(\mathbf{z}_{k,i})) \mathbf{w}_k}_{\rm pulling \ part}  \nonumber\\
    &+ \underbrace{\sum_{k' \ne k}^K (p_{k'}^{(b)}(\mathbf{m}_{i,b}) + p_{k'}^{(l)}(\mathbf{z}_{k,i}))\mathbf{w}_{k'}}_{\rm forcing \ part} 
\label{eq:wrt_feat}
\end{align}

In the gradient \textit{w.r.t} features, the \textit{pulling part} directs the feature $\widetilde{\mathbf{z}}_{k,i}$ towards the weight of its class $\mathbf{w}_c$, and the \textit{forcing part} repels it from wrong classes. Regarding the weight factors, the probability of bias attribute $p_k^{(b)}$ also re-weights the influence of classifier weights. Bias-aligned samples, with high $p_k^{(b)}$ probability, will have smaller pulling effects towards the classifier weight $\mathbf{w}_k$, which avoids the dominance of bias-aligned features around the weight centers and hinders the tendency of shortcut learning. In comparison, the bias-conflicting samples are granted stronger pulling and pushing effects, which strengthens their convergence toward the right class. The detailed theoretical justification of our method, along with the comparison with vanilla training, are available in Appendix \ref{sup:theo}.
\section{Experiments}
\begin{table*}[bht]
\caption{Comparison of debiasing performance on synthetic datasets. We report the accuracy on the unbiased test sets of Colored MNIST and Corrupted CIFAR-10. Best performances are marked in bold, and the number in brackets indicates the improvement compared to the best result in baselines. $(^{\ast})$ and $(^{\diamond})$ denote methods with/without bias supervision respectively.}
\label{table:syn}
\centering
\newcommand{\stdv}[1]{\scriptsize$\pm$#1}
\newcommand{\improv}[1]{\footnotesize\textcolor[RGB]{73,84,138}{\textbf{(+#1)}}}
\newcommand{\reduce}[1]{\footnotesize\textcolor[RGB]{81,69,73}{\textbf{(-#1)}}}
\begin{adjustbox}{width=\textwidth}
    \begin{tabular}{cccccccccl}
    \toprule[1.5pt]
    \makebox[0.035\textwidth][c]{Dataset} & \makebox[0.03\textwidth][c]{Ratio(\%)} & \makebox[0.05\textwidth][c]{Vanilla} &
\makebox[0.05\textwidth][c]{LfF$^{\diamond}$\cite{nam2020learning}} &
    \makebox[0.05\textwidth][c]{LfF+BE$^{\diamond}$\cite{lee2023revisiting}} &
    \makebox[0.05\textwidth][c]{EnD$^{\ast}$\cite{tartaglione2021end}} &
    \makebox[0.05\textwidth][c]{SD$^{\ast}$\cite{zhang2023benign}} &
    \makebox[0.05\textwidth][c]{DisEnt$^{\ast}$\cite{lee2021learning}} &
    \makebox[0.09\textwidth][c]{Selecmix$^{\diamond}$\cite{hwang2022selecmix}} &
    \makebox[0.12\textwidth][c]{\textbf{ETF-Debias}} \\
    \midrule
    \multirow{4}{*}{\makecell{Colored \\ MNIST}}
    & 0.5 & {32.22}\stdv{0.13} & 57.78\stdv{0.81} & {69.69}\stdv{1.99} & {35.93}\stdv{0.40} & {56.96}\stdv{0.37} & {68.83}\stdv{1.62} & {70.53}\stdv{0.46} & \textbf{71.63}\stdv{0.28} \improv{1.10} \\ 
    & 1.0 & {48.45}\stdv{0.06} & 72.29\stdv{1.69} & {80.90}\stdv{1.40} & {49.32}\stdv{0.58} & {72.46}\stdv{0.18} & {79.49}\stdv{1.44} & \textbf{83.34}\stdv{0.37} & {81.97}\stdv{0.26} \reduce{1.37} \\ 
    & 2.0 & {58.90}\stdv{0.12} & 79.51\stdv{1.82} & {84.90}\stdv{1.14} & {65.58}\stdv{0.46} & {79.37}\stdv{0.46} & {84.56}\stdv{1.19} & {85.90}\stdv{0.23} & \textbf{86.00}\stdv{0.03} \improv{0.10} \\ 
    & 5.0 & {74.19}\stdv{0.04} & 83.96\stdv{1.44} & {90.28}\stdv{0.18} & {80.70}\stdv{0.17} & {88.89}\stdv{0.21} & {88.83}\stdv{0.15} & {91.27}\stdv{0.31} & \textbf{91.36}\stdv{0.21} \improv{0.09} \\ 
    \midrule
    \multirow{4}{*}{\makecell{Corrupted \\ CIFAR-10}} 
    & 0.5 & {17.06}\stdv{0.12} & 31.00\stdv{2.67} & {23.68}\stdv{0.50} & {14.30}\stdv{0.10} & {36.66}\stdv{0.74} & {30.12}\stdv{1.60} & {33.30}\stdv{0.26} & \textbf{40.06}\stdv{0.03} \improv{3.40} \\ 
    & 1.0 & {21.48}\stdv{0.55} & 34.33\stdv{1.76} & {30.72}\stdv{0.12} & {20.17}\stdv{0.19} & {45.66}\stdv{1.05} & {35.28}\stdv{1.39} & {38.72}\stdv{0.27} & \textbf{47.52}\stdv{0.26} \improv{1.86} \\ 
    & 2.0 & {27.15}\stdv{0.46} & 39.68\stdv{1.15} & {42.22}\stdv{0.60} & {30.10}\stdv{0.54} & {50.11}\stdv{0.69} & {40.34}\stdv{1.41} & {47.09}\stdv{0.17} & \textbf{54.64}\stdv{0.42} \improv{4.53}\\ 
    & 5.0 & {39.46}\stdv{0.58} & 53.04\stdv{0.76} & {57.93}\stdv{0.58} & {45.85}\stdv{0.21} & {62.43}\stdv{0.57} & {49.99}\stdv{0.84} & {54.69}\stdv{0.29} & \textbf{65.34}\stdv{0.60} \improv{2.91} \\ 
    \bottomrule[1.5pt]
    \end{tabular}
    \end{adjustbox}
\end{table*}
\begin{table*}[bht]
\caption{Comparison of debiasing performance on real-world datasets. We report the accuracy on the unbiased test sets of Biased FFHQ, Dogs \& Cats, and BAR. The class-wise accuracy on BAR is reported in Appendix \ref{sup:bar}. Best performances are marked in bold, and the number in brackets indicates the improvement compared to the best result in baselines. $(^{\ast})$ and $(^{\diamond})$ denote methods with/without bias supervision respectively.}
\label{table:rea}
\centering
\newcommand{\stdv}[1]{\scriptsize$\pm$#1}
\newcommand{\improv}[1]{\footnotesize\textcolor[RGB]{73,84,138}{\textbf{(+#1)}}}
\newcommand{\reduce}[1]{\footnotesize\textcolor[RGB]{81,69,73}{\textbf{(-#1)}}}
\begin{adjustbox}{width=\textwidth}
\begin{tabular}{cccccccccl}
\toprule[1.5pt]
    \makebox[0.035\textwidth][c]{Dataset} & \makebox[0.03\textwidth][c]{Ratio(\%)} & \makebox[0.05\textwidth][c]{Vanilla} &
\makebox[0.05\textwidth][c]{LfF$^{\diamond}$\cite{nam2020learning}} &
    \makebox[0.05\textwidth][c]{LfF+BE$^{\diamond}$\cite{lee2023revisiting}} &
    \makebox[0.05\textwidth][c]{EnD$^{\ast}$\cite{tartaglione2021end}} &
    \makebox[0.05\textwidth][c]{SD$^{\ast}$\cite{zhang2023benign}} &
    \makebox[0.05\textwidth][c]{DisEnt$^{\ast}$\cite{lee2021learning}} &
    \makebox[0.09\textwidth][c]{Selecmix$^{\diamond}$\cite{hwang2022selecmix}} &
    \makebox[0.12\textwidth][c]{\textbf{ETF-Debias}} \\
\midrule
\multirow{4}{*}{\makecell{Biased \\ FFHQ}}
& 0.5 & {53.27}\stdv{0.61} & 65.60\stdv{2.27} & {67.07}\stdv{2.37} & {55.93}\stdv{1.62} & {65.60}\stdv{0.20} & {63.07}\stdv{1.14} & {65.00}\stdv{0.82} & \textbf{73.60}\stdv{1.22} \improv{6.53} \\ 
& 1.0 & {57.13}\stdv{0.64} & 72.33\stdv{2.19} & {73.53}\stdv{1.62} & {61.13}\stdv{0.50} & {69.20}\stdv{0.20} & {68.53}\stdv{2.32} & {67.50}\stdv{0.30} & \textbf{76.53}\stdv{1.10} \improv{3.00}\\ 
& 2.0 & {67.67}\stdv{0.81} & 74.80\stdv{2.03} & {80.20}\stdv{2.78} & {66.87}\stdv{0.64} & {78.40}\stdv{0.20} & {72.00}\stdv{2.51} & {69.80}\stdv{0.87} & \textbf{85.20}\stdv{0.61} \improv{5.00} \\ 
& 5.0 & {78.87}\stdv{0.83} & 80.27\stdv{2.02} & {87.40}\stdv{2.00} & {80.87}\stdv{0.42} & {84.80}\stdv{0.20} & {80.60}\stdv{0.53} & {83.47}\stdv{0.61} & \textbf{94.00}\stdv{0.72} \improv{6.60}\\ 
\midrule
\multirow{2}{*}{\makecell{Dogs \& Cats}} 
& 1.0 & {51.96}\stdv{0.90} & 71.17\stdv{5.24} & {78.87}\stdv{2.40} & {51.91}\stdv{0.24} & {78.13}\stdv{1.06} & {65.13}\stdv{2.07} & {54.19}\stdv{1.61} & \textbf{80.07}\stdv{0.90} \improv{1.20}\\ 
& 5.0 & {76.59}\stdv{1.27} & 85.83\stdv{1.62} & {88.60}\stdv{1.21} & {79.07}\stdv{0.28} & {89.12}\stdv{0.18} & {82.47}\stdv{2.86} & {81.50}\stdv{1.06} & \textbf{92.18}\stdv{0.62} \improv{3.06} \\ 
\midrule
\multirow{2}{*}{\makecell{BAR}}
& 1.0 & {68.00}\stdv{0.43} & 68.30\stdv{0.97} & {71.70}\stdv{1.33} & {68.25}\stdv{0.19} & {67.33}\stdv{0.35} & {69.30}\stdv{1.27} & {69.83}\stdv{1.02} & \textbf{72.79}\stdv{0.21} \improv{1.09}\\ 
& 5.0 & {79.34}\stdv{0.19} & 80.25\stdv{1.27} & {82.00}\stdv{1.24} & {78.86}\stdv{0.36} & {79.10}\stdv{0.42} & {81.19}\stdv{0.70} & {78.79}\stdv{0.52} & \textbf{83.66}\stdv{0.21} \improv{1.66}\\ 
\bottomrule[1.5pt]
\end{tabular}
\end{adjustbox}
\end{table*}

\subsection{Experimental Settings}
\noindent \textbf{Datasets and models.}
We validate the effectiveness of ETF-Debias on general debiasing benchmarks, which cover various types of bias attributes including color, corruption, gender, and background. We adopt 2 synthetic biased datasets, Colored MNIST \cite{kim2019learning} and Corrupted CIFAR-10 \cite{hendrycks2018benchmarking} with the ratio of bias-conflicting training samples \{0.5\%, 1.0\%, 2.0\%, 5.0\%\}, and 3 real-world biased datasets, Biased FFHQ (BFFHQ) \cite{lee2021learning} with bias ratio \{0.5\%, 1.0\%, 2.0\%, 5.0\%\}, BAR \cite{nam2020learning}, and Dogs \& Cats \cite{kim2019learning} with bias ratio \{1.0\%, 5.0\%\}.

\noindent As for the model architecture, we adopt a three-layer MLP for Colored MNIST and ResNet-20 \cite{he2016deep} for other datasets. Since BAR has a tiny training set, we follow the previous work \cite{lee2023revisiting} and initialize the parameters with pre-trained models on corresponding datasets. All results are averaged over three independent trials. More details about datasets and implementation are available in Appendix \ref{sup:data}.

\noindent \textbf{Baselines.}
According to the three categories of debiased learning in Section \ref{sec:2}, we compare the performance of ETF-Debias with six recent methods. For reweight-based debiasing, we consider LfF \cite{nam2020learning} with auxiliary bias models, and its improved version LfF+BE \cite{lee2023revisiting}. For disentangle-based debiasing, we consider EnD \cite{tartaglione2021end} and SD \cite{zhang2023benign}, which stem from the Information Bottleneck theory and causal intervention respectively. For augmentation-based debiasing, we consider DisEnt \cite{lee2021learning} and Selecmix \cite{hwang2022selecmix}, to include both the feature-level and image-level augmentations.

\subsection{Main Results}
\noindent \textbf{Comparison on synthetic datasets.}
To display the debiasing performance, we report the accuracy on the unbiased test set of 2 synthetic datasets in Tab. \ref{table:syn}. It’s notable that ETF-Debias consistently outperforms baselines in the generalization capability towards test samples, on almost all levels of bias ratio. We observe that some baseline methods (e.g., EnD) do not display a satisfactory debiasing effect on synthetic datasets, as they rely heavily on diverse contrastive samples to identify and mitigate the bias features. In contrast, our approach directly provides the approximated shortcut features as training primes, which achieves superior performance on synthetic bias attributes. 

\noindent \textbf{Comparison on real-world datasets.} To verify the scalability of ETF-Debias in real-world scenarios with more diverse bias attributes, we test our method on 3 real-world biased datasets in Tab. \ref{table:rea}. We observe that ETF-Debias shows an even greater performance gain on real-world datasets than on synthetic ones, which may be attributed to the semantically meaningful prime features constructed with the simplex ETF structure. On the large-scale BFFHQ dataset, our method achieves up to 6.6\% accuracy improvements compared to baseline methods, demonstrating its potential in real-world applications.  

\noindent \textbf{Convergence of Neural Collapse.} In Fig. \ref{fig:NC}, we display the trajectory of NC metrics during training on the Corrupted CIFAR-10 dataset. Guided by the prime features, the model establishes a right correlation and shows a much better convergence property on biased datasets, contributing to the superior generalization capability. More convergence results are available in Appendix \ref{sup:converge}. 

\subsection{Ablation Study}
\label{sec:5.3}
\begin{figure}[htbp]
    \centering
\includegraphics[width=\columnwidth]{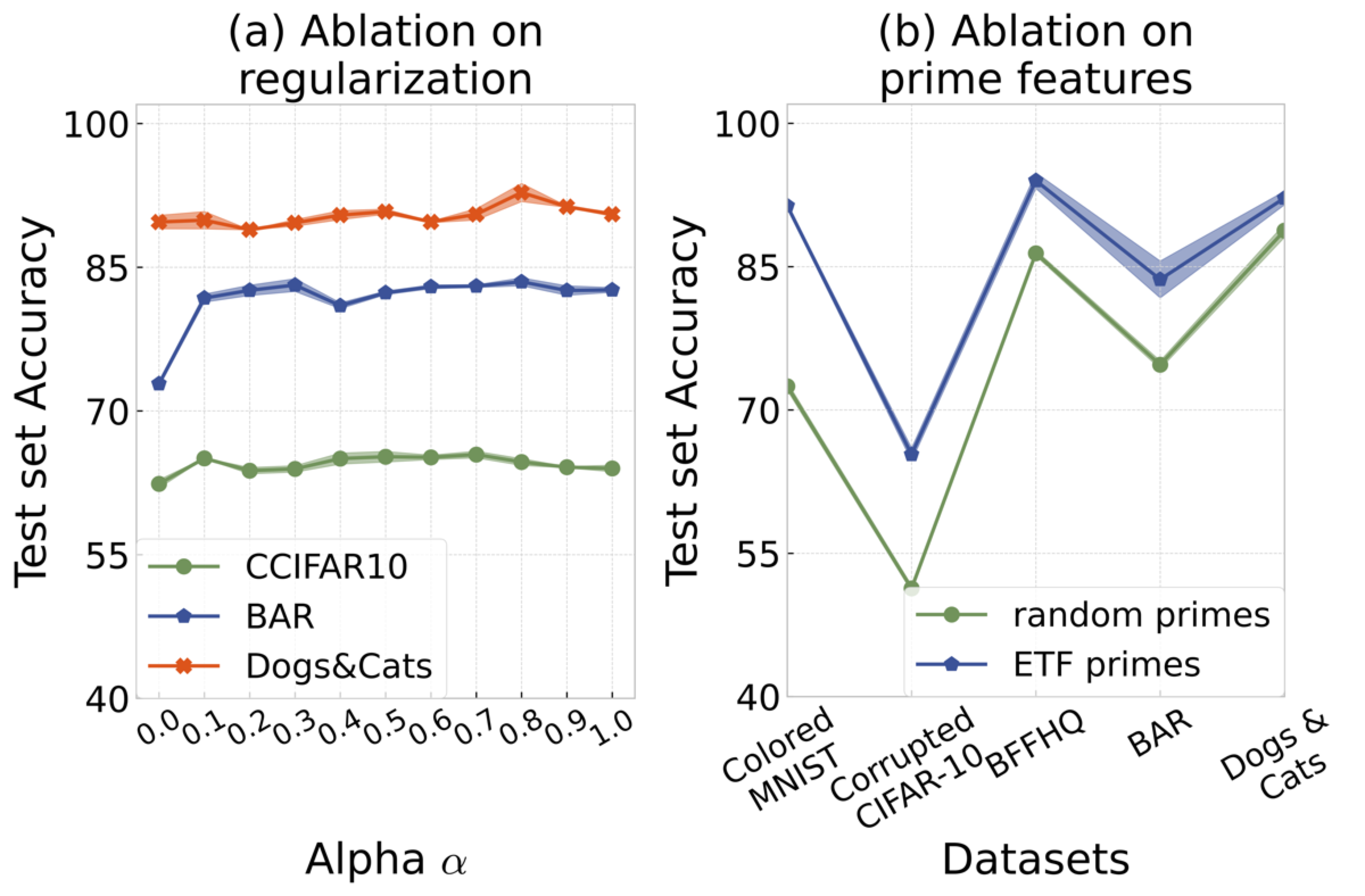}
\caption{Ablation studies on hyper-parameter and prime features. We report (a) test set accuracy on different datasets, with hyper-parameter $\alpha$ ranging from 0.0 to 1.0, and (b) test set accuracy on 5 datasets with different prime features. The shaded areas represent the standard deviation, and the bias ratio of all datasets is 5\%.}
\label{fig:ablation}
\vspace{-1em}
\end{figure}

\noindent \textbf{Ablation on the influence of regularization. } To measure the sensitivity of our method to different levels of prime reinforcement regularization, we compare the accuracy on the unbiased test set with $\alpha$ range from 0.0 to 1.0 in Fig. \ref{fig:ablation}(a). It's been shown that the debiasing performance remains significant with different strengths of regularization, and achieves extra performance gain with the proper level of prime reinforcement.

\noindent \textbf{Ablation on the influence of ETF prime features.} As illustrated before, we choose the vertices of ETF as the ``perfectly learned" shortcut features, thus redirecting the model's attention to intrinsic correlations. To demonstrate the efficacy of ETF prime features, we compare the results of randomly initialized prime features with the same dimension as the ETF-based ones. As shown in Fig. \ref{fig:ablation}(b), the randomly initialized primes suffer a severe performance degradation, underscoring the advantages of ETF-based prime features in approximating the optimal structure. 

\subsection{Visualization}
\label{sec:5.4}
\begin{figure}[htbp]
    \centering
\includegraphics[width=\columnwidth]{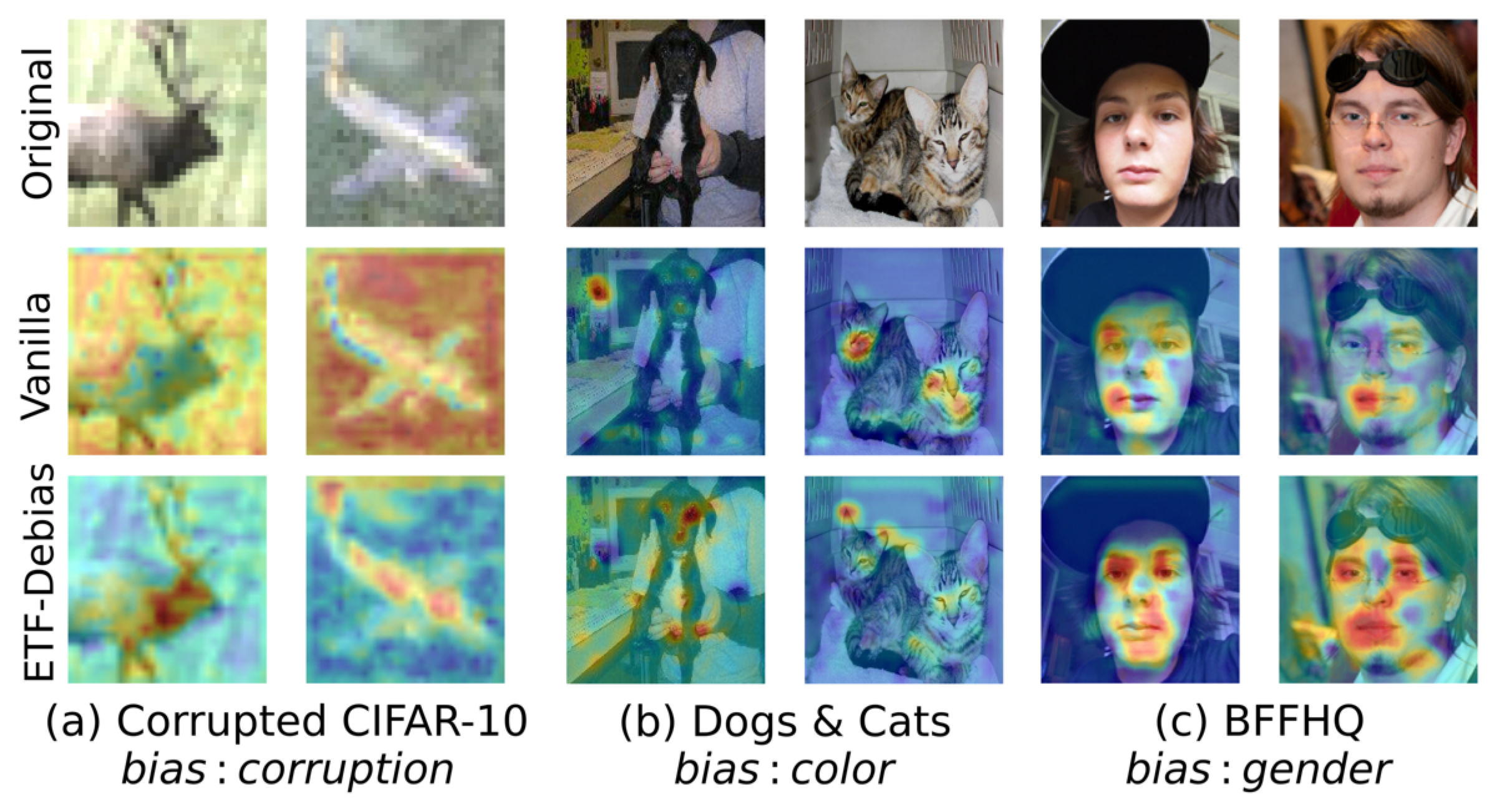}
\caption{The comparison of visualization results on bias-conflicting samples (first row) between vanilla models (second row) and ETF-Debias models (third row). We display the result of CAM \cite{zhou2016learning} on (a) Corrupted CIFAR-10 dataset, with the bias attribute as different types of corruption on the entire image, (b) Dogs \& Cats dataset, with the bias attribute as the color of animals and (c) BFFHQ dataset, with the bias attribute as gender.}
\vspace{-0.5em}
\label{fig:vis}
\end{figure}
To intuitively reveal the effectiveness of our method, we compare the CAM \cite{zhou2016learning} visualization results on vanilla models and debiased models trained with ETF-Debias. As shown in Fig. \ref{fig:vis}, the model's attention is significantly rectified with ETF-Debias. For example, on Corrupted CIFAR-10 dataset, vanilla models are easily misled by the corruptions on the entire images, but with the guide of prime features in our method, the debiased models shift their attention to the objects themselves. It's also notable that our method circumvents the wrong attention area in classification and encourages the focus on more discriminative and finer-grained regions. On datasets of facial recognition, our method also breaks the spurious correlation on specific visual attributes \cite{lee2023revisiting} and considers more facial features, as shown in Fig. \ref{fig:vis}(c). More visualization results are available in Appendix \ref{sup:vis}.
\section{Conclusion}
In this paper, we propose an avoid-shortcut learning framework with the insights of the Neural Collapse phenomenon. By extending the analysis of Neural Collapse to biased datasets, we introduce the simplex ETF as the prime features to redirect the model's attention to intrinsic correlations. With the state-of-the-art debiasing performance on various benchmarks, we hope our work may advance the understanding of Neural Collapse and shed light on the fundamental solutions to model debiasing.
\section*{Acknowledgement}
We appreciate the valuable comments from the anonymous reviewers that improves the paper's quality. This work was supported in part by the National Key Research and Development Program (2021YFB3101200), National
Natural Science Foundation of China (U1736208, U1836210, U1836213, 62172104, 62172105, 61902374, 62102093, 62102091). Min Yang is a faculty of Shanghai Institute of Intelligent Electronics \& Systems, Shanghai Insitute for Advanced Communication and Data Science, and Engineering Research Center of Cyber Security Auditing and Monitoring, Ministry of Education, China. 
{
    \small
\bibliographystyle{ieeenat_fullname}
    \bibliography{main}

\begin{thebibliography}{46}
\providecommand{\natexlab}[1]{#1}
\providecommand{\url}[1]{\texttt{#1}}
\expandafter\ifx\csname urlstyle\endcsname\relax
  \providecommand{\doi}[1]{doi: #1}\else
  \providecommand{\doi}{doi: \begingroup \urlstyle{rm}\Url}\fi

\bibitem[Balayn and G{\"u}rses(2021)]{balayn2021beyond}
Agathe Balayn and Seda G{\"u}rses.
\newblock Beyond debiasing: Regulating ai and its inequalities.
\newblock \emph{EDRi Report. https://edri. org/wp-content/uploads/2021/09/EDRi\_Beyond-Debiasing-Report\_Online. pdf}, 2021.

\bibitem[Dang et~al.(2023)Dang, Nguyen, Tran, Tran, and Ho]{dang2023neural}
Hien Dang, Tan Nguyen, Tho Tran, Hung Tran, and Nhat Ho.
\newblock Neural collapse in deep linear network: From balanced to imbalanced data.
\newblock \emph{ICML}, 2023.

\bibitem[Deviyani(2022)]{deviyani2022assessing}
Athiya Deviyani.
\newblock Assessing dataset bias in computer vision.
\newblock \emph{arXiv preprint arXiv:2205.01811}, 2022.

\bibitem[Fan et~al.(2022)Fan, Wang, Mo, Shi, and Tang]{fan2022debiasing}
Shaohua Fan, Xiao Wang, Yanhu Mo, Chuan Shi, and Jian Tang.
\newblock Debiasing graph neural networks via learning disentangled causal substructure.
\newblock \emph{NeurIPS}, 35:\penalty0 24934--24946, 2022.

\bibitem[Fang et~al.(2021)Fang, He, Long, and Su]{fang2021exploring}
Cong Fang, Hangfeng He, Qi Long, and Weijie~J Su.
\newblock Exploring deep neural networks via layer-peeled model: Minority collapse in imbalanced training.
\newblock \emph{Proceedings of the National Academy of Sciences}, 118\penalty0 (43):\penalty0 e2103091118, 2021.

\bibitem[Galanti et~al.(2022)Galanti, Gy{\"o}rgy, and Hutter]{galanti2022role}
Tomer Galanti, Andr{\'a}s Gy{\"o}rgy, and Marcus Hutter.
\newblock On the role of neural collapse in transfer learning.
\newblock In \emph{ICLR}, 2022.

\bibitem[Guo et~al.(2022)Guo, Yang, and Abbasi]{guo2022auto}
Yue Guo, Yi Yang, and Ahmed Abbasi.
\newblock Auto-debias: Debiasing masked language models with automated biased prompts.
\newblock In \emph{ACL}, pages 1012--1023, 2022.

\bibitem[Han et~al.(2022)Han, Papyan, and Donoho]{han2021neural}
XY Han, Vardan Papyan, and David~L Donoho.
\newblock Neural collapse under mse loss: Proximity to and dynamics on the central path.
\newblock In \emph{ICLR}, 2022.

\bibitem[He et~al.(2016)He, Zhang, Ren, and Sun]{he2016deep}
Kaiming He, Xiangyu Zhang, Shaoqing Ren, and Jian Sun.
\newblock Deep residual learning for image recognition.
\newblock In \emph{CVPR}, pages 770--778, 2016.

\bibitem[Hendrycks and Dietterich(2018)]{hendrycks2018benchmarking}
Dan Hendrycks and Thomas Dietterich.
\newblock Benchmarking neural network robustness to common corruptions and perturbations.
\newblock In \emph{ICLR}, 2018.

\bibitem[Hirota et~al.(2023)Hirota, Nakashima, and Garcia]{hirota2023model}
Yusuke Hirota, Yuta Nakashima, and Noa Garcia.
\newblock Model-agnostic gender debiased image captioning.
\newblock In \emph{CVPR}, pages 15191--15200, 2023.

\bibitem[Huang et~al.(2008)Huang, Mattar, Berg, and Learned-Miller]{huang2008labeled}
Gary~B Huang, Marwan Mattar, Tamara Berg, and Eric Learned-Miller.
\newblock Labeled faces in the wild: A database forstudying face recognition in unconstrained environments.
\newblock In \emph{Workshop on faces in'Real-Life'Images: detection, alignment, and recognition}, 2008.

\bibitem[Hwang et~al.(2022)Hwang, Lee, Kwak, Oh, Teney, Kim, and Zhang]{hwang2022selecmix}
Inwoo Hwang, Sangjun Lee, Yunhyeok Kwak, Seong~Joon Oh, Damien Teney, Jin-Hwa Kim, and Byoung-Tak Zhang.
\newblock Selecmix: Debiased learning by contradicting-pair sampling.
\newblock \emph{NeurIPS}, 35:\penalty0 14345--14357, 2022.

\bibitem[Karras et~al.(2019)Karras, Laine, and Aila]{karras2019style}
Tero Karras, Samuli Laine, and Timo Aila.
\newblock A style-based generator architecture for generative adversarial networks.
\newblock In \emph{CVPR}, pages 4401--4410, 2019.

\bibitem[Kim et~al.(2019)Kim, Kim, Kim, Kim, and Kim]{kim2019learning}
Byungju Kim, Hyunwoo Kim, Kyungsu Kim, Sungjin Kim, and Junmo Kim.
\newblock Learning not to learn: Training deep neural networks with biased data.
\newblock In \emph{CVPR}, pages 9012--9020, 2019.

\bibitem[Kim et~al.(2022)Kim, Hwang, Ahn, Park, and Kwak]{kim2022learning}
Nayeong Kim, Sehyun Hwang, Sungsoo Ahn, Jaesik Park, and Suha Kwak.
\newblock Learning debiased classifier with biased committee.
\newblock \emph{NeurIPS}, 35:\penalty0 18403--18415, 2022.

\bibitem[Krizhevsky et~al.(2009)Krizhevsky, Hinton, et~al.]{krizhevsky2009learning}
Alex Krizhevsky, Geoffrey Hinton, et~al.
\newblock Learning multiple layers of features from tiny images.
\newblock 2009.

\bibitem[Lee et~al.(2021)Lee, Kim, Lee, Lee, and Choo]{lee2021learning}
Jungsoo Lee, Eungyeup Kim, Juyoung Lee, Jihyeon Lee, and Jaegul Choo.
\newblock Learning debiased representation via disentangled feature augmentation.
\newblock \emph{NeurIPS}, 34:\penalty0 25123--25133, 2021.

\bibitem[Lee et~al.(2023)Lee, Park, Kim, Lee, Choi, and Choo]{lee2023revisiting}
Jungsoo Lee, Jeonghoon Park, Daeyoung Kim, Juyoung Lee, Edward Choi, and Jaegul Choo.
\newblock Revisiting the importance of amplifying bias for debiasing.
\newblock In \emph{AAAI}, pages 14974--14981, 2023.

\bibitem[Li et~al.(2023)Li, Shang, He, Lin, and Wu]{li2023no}
Zexi Li, Xinyi Shang, Rui He, Tao Lin, and Chao Wu.
\newblock No fear of classifier biases: Neural collapse inspired federated learning with synthetic and fixed classifier.
\newblock \emph{ICCV}, 2023.

\bibitem[Lim et~al.(2023)Lim, Kim, Kim, Ahn, Shin, Yang, and Han]{lim2023biasadv}
Jongin Lim, Youngdong Kim, Byungjai Kim, Chanho Ahn, Jinwoo Shin, Eunho Yang, and Seungju Han.
\newblock Biasadv: Bias-adversarial augmentation for model debiasing.
\newblock In \emph{CVPR}, pages 3832--3841, 2023.

\bibitem[Lyu et~al.(2023)Lyu, Li, Yang, de~Rijke, Ren, Zhao, Yin, and Ren]{lyu2023feature}
Yougang Lyu, Piji Li, Yechang Yang, Maarten de Rijke, Pengjie Ren, Yukun Zhao, Dawei Yin, and Zhaochun Ren.
\newblock Feature-level debiased natural language understanding.
\newblock In \emph{AAAI}, pages 13353--13361, 2023.

\bibitem[Nam et~al.(2020)Nam, Cha, Ahn, Lee, and Shin]{nam2020learning}
Junhyun Nam, Hyuntak Cha, Sungsoo Ahn, Jaeho Lee, and Jinwoo Shin.
\newblock Learning from failure: De-biasing classifier from biased classifier.
\newblock \emph{NeurIPS}, 33:\penalty0 20673--20684, 2020.

\bibitem[Papyan et~al.(2020)Papyan, Han, and Donoho]{papyan2020prevalence}
Vardan Papyan, XY Han, and David~L Donoho.
\newblock Prevalence of neural collapse during the terminal phase of deep learning training.
\newblock \emph{Proceedings of the National Academy of Sciences}, 117\penalty0 (40):\penalty0 24652--24663, 2020.

\bibitem[Peifeng et~al.(2023)Peifeng, Xu, Wen, Yang, Shao, and Huang]{peifeng2023feature}
Gao Peifeng, Qianqian Xu, Peisong Wen, Zhiyong Yang, Huiyang Shao, and Qingming Huang.
\newblock Feature directions matter: Long-tailed learning via rotated balanced representation.
\newblock \emph{ICML}, 2023.

\bibitem[Rangamani et~al.(2023)Rangamani, Lindegaard, Galanti, and Poggio]{rangamani2023feature}
Akshay Rangamani, Marius Lindegaard, Tomer Galanti, and Tomaso~A Poggio.
\newblock Feature learning in deep classifiers through intermediate neural collapse.
\newblock In \emph{ICML}, pages 28729--28745. PMLR, 2023.

\bibitem[Robinson et~al.(2021)Robinson, Sun, Yu, Batmanghelich, Jegelka, and Sra]{robinson2021can}
Joshua Robinson, Li Sun, Ke Yu, Kayhan Batmanghelich, Stefanie Jegelka, and Suvrit Sra.
\newblock Can contrastive learning avoid shortcut solutions?
\newblock \emph{NeurIPS}, 34:\penalty0 4974--4986, 2021.

\bibitem[Saranrittichai et~al.(2022)Saranrittichai, Mummadi, Blaiotta, Munoz, and Fischer]{saranrittichai2022overcoming}
Piyapat Saranrittichai, Chaithanya~Kumar Mummadi, Claudia Blaiotta, Mauricio Munoz, and Volker Fischer.
\newblock Overcoming shortcut learning in a target domain by generalizing basic visual factors from a source domain.
\newblock In \emph{ECCV}, pages 294--309. Springer, 2022.

\bibitem[Shinoda et~al.(2023)Shinoda, Sugawara, and Aizawa]{shinoda2023shortcut}
Kazutoshi Shinoda, Saku Sugawara, and Akiko Aizawa.
\newblock Which shortcut solution do question answering models prefer to learn?
\newblock In \emph{AAAI}, pages 13564--13572, 2023.

\bibitem[Tartaglione et~al.(2021)Tartaglione, Barbano, and Grangetto]{tartaglione2021end}
Enzo Tartaglione, Carlo~Alberto Barbano, and Marco Grangetto.
\newblock End: Entangling and disentangling deep representations for bias correction.
\newblock In \emph{CVPR}, pages 13508--13517, 2021.

\bibitem[Thrampoulidis et~al.(2022)Thrampoulidis, Kini, Vakilian, and Behnia]{thrampoulidis2022imbalance}
Christos Thrampoulidis, Ganesh~Ramachandra Kini, Vala Vakilian, and Tina Behnia.
\newblock Imbalance trouble: Revisiting neural-collapse geometry.
\newblock \emph{Advances in Neural Information Processing Systems}, 35:\penalty0 27225--27238, 2022.

\bibitem[Wang et~al.(2021)Wang, Zhou, Sun, and Zhang]{wang2021causal}
Tan Wang, Chang Zhou, Qianru Sun, and Hanwang Zhang.
\newblock Causal attention for unbiased visual recognition.
\newblock In \emph{CVPR}, pages 3091--3100, 2021.

\bibitem[Wen et~al.(2022)Wen, Qian, Lin, Teng, Jayaraman, and Gao]{wen2022fighting}
Chuan Wen, Jianing Qian, Jierui Lin, Jiaye Teng, Dinesh Jayaraman, and Yang Gao.
\newblock Fighting fire with fire: Avoiding dnn shortcuts through priming.
\newblock In \emph{ICML}, pages 23723--23750. PMLR, 2022.

\bibitem[Wen et~al.(2021)Wen, Xu, Tan, Wu, and Wu]{wen2021debiased}
Zhiquan Wen, Guanghui Xu, Mingkui Tan, Qingyao Wu, and Qi Wu.
\newblock Debiased visual question answering from feature and sample perspectives.
\newblock \emph{NeurIPS}, 34:\penalty0 3784--3796, 2021.

\bibitem[Xie et~al.(2023)Xie, Yang, Cai, and He]{xie2023neural}
Liang Xie, Yibo Yang, Deng Cai, and Xiaofei He.
\newblock Neural collapse inspired attraction--repulsion-balanced loss for imbalanced learning.
\newblock \emph{Neurocomputing}, 527:\penalty0 60--70, 2023.

\bibitem[Yang et~al.(2022{\natexlab{a}})Yang, Chen, Li, Xie, Lin, and Tao]{yang2022inducing}
Yibo Yang, Shixiang Chen, Xiangtai Li, Liang Xie, Zhouchen Lin, and Dacheng Tao.
\newblock Inducing neural collapse in imbalanced learning: Do we really need a learnable classifier at the end of deep neural network?
\newblock \emph{NeurIPS}, 35:\penalty0 37991--38002, 2022{\natexlab{a}}.

\bibitem[Yang et~al.(2022{\natexlab{b}})Yang, Yuan, Li, Lin, Torr, and Tao]{yang2022neural}
Yibo Yang, Haobo Yuan, Xiangtai Li, Zhouchen Lin, Philip Torr, and Dacheng Tao.
\newblock Neural collapse inspired feature-classifier alignment for few-shot class-incremental learning.
\newblock In \emph{ICLR}, 2022{\natexlab{b}}.

\bibitem[Yang et~al.(2023)Yang, Steinhardt, and Hu]{yang2023neurons}
Yongyi Yang, Jacob Steinhardt, and Wei Hu.
\newblock Are neurons actually collapsed? on the fine-grained structure in neural representations.
\newblock \emph{ICML}, 2023.

\bibitem[Yann~LeCun(2010)]{mnist}
Corinna Cortes~and Yann~LeCun.
\newblock Mnist handwritten digit database.
\newblock \emph{Available at \url{http://yann.lecun.com/exdb/mnist/}}, 2010.

\bibitem[Yaras et~al.(2022)Yaras, Wang, Zhu, Balzano, and Qu]{yaras2022neural}
Can Yaras, Peng Wang, Zhihui Zhu, Laura Balzano, and Qing Qu.
\newblock Neural collapse with normalized features: A geometric analysis over the riemannian manifold.
\newblock \emph{NeurIPS}, 35:\penalty0 11547--11560, 2022.

\bibitem[Zhang et~al.(2023{\natexlab{a}})Zhang, Zhang, Liu, Wang, Gao, Zhang, and Guo]{zhang2023debiasing}
Qing Zhang, Xiaoying Zhang, Yang Liu, Hongning Wang, Min Gao, Jiheng Zhang, and Ruocheng Guo.
\newblock Debiasing recommendation by learning identifiable latent confounders.
\newblock \emph{KDD}, 2023{\natexlab{a}}.

\bibitem[Zhang et~al.(2023{\natexlab{b}})Zhang, Sang, Wang, Jiang, and Wang]{zhang2023benign}
Yi Zhang, Jitao Sang, Junyang Wang, Dongmei Jiang, and Yaowei Wang.
\newblock Benign shortcut for debiasing: Fair visual recognition via intervention with shortcut features.
\newblock \emph{ACM MM}, 2023{\natexlab{b}}.

\bibitem[Zhong et~al.(2023)Zhong, Cui, Yang, Wu, Qi, Zhang, and Jia]{zhong2023understanding}
Zhisheng Zhong, Jiequan Cui, Yibo Yang, Xiaoyang Wu, Xiaojuan Qi, Xiangyu Zhang, and Jiaya Jia.
\newblock Understanding imbalanced semantic segmentation through neural collapse.
\newblock In \emph{CVPR}, pages 19550--19560, 2023.

\bibitem[Zhou et~al.(2016)Zhou, Khosla, Lapedriza, Oliva, and Torralba]{zhou2016learning}
Bolei Zhou, Aditya Khosla, Agata Lapedriza, Aude Oliva, and Antonio Torralba.
\newblock Learning deep features for discriminative localization.
\newblock In \emph{CVPR}, pages 2921--2929, 2016.

\bibitem[Zhou et~al.(2022)Zhou, You, Li, Liu, Liu, Qu, and Zhu]{zhou2022all}
Jinxin Zhou, Chong You, Xiao Li, Kangning Liu, Sheng Liu, Qing Qu, and Zhihui Zhu.
\newblock Are all losses created equal: A neural collapse perspective.
\newblock \emph{NeurIPS}, 35:\penalty0 31697--31710, 2022.

\bibitem[Zhu et~al.(2021)Zhu, Ding, Zhou, Li, You, Sulam, and Qu]{zhu2021geometric}
Zhihui Zhu, Tianyu Ding, Jinxin Zhou, Xiao Li, Chong You, Jeremias Sulam, and Qing Qu.
\newblock A geometric analysis of neural collapse with unconstrained features.
\newblock \emph{NeurIPS}, 34:\penalty0 29820--29834, 2021.

\end{thebibliography}
}

\appendix
\maketitlesupplementary

\section{Detailed Theoretical Justification}
\label{sup:theo}

\subsection{Analysis of Vanilla Training}
To illustrate why vanilla models tend to pursue shortcut learning, we follow the analysis of previous works \cite{yang2022inducing, zhong2023understanding} and re-examine the issue of biased classification from the perspective of gradients. 

Following the definition in Section \ref{sec:3.1}, we denote $\mathbf{x}_{k,i}$ as the $i$-th sample of the $k$-th class,  $\mathbf{z}_{k,i} \in \mathbb{R}^d$ as its corresponding last-layer feature, and $\mathbf{W}=[\mathbf{w}_1, ..., \mathbf{w}_K]$ as the weights of classifier. In vanilla training, the cross-entropy loss is defined as: 
\renewcommand{\theequation}{A.\arabic{equation}}
\setcounter{equation}{0}
\begin{equation}
    \mathcal{L}_{\rm CE}(\mathbf{z}_{k,i}, \mathbf{W})=
    -\log \left( \frac{\exp (\mathbf{z}_{k,i}^\mathrm{T} \mathbf{w}_k)}{\sum_{k'=1}^K \exp(\mathbf{z}_{k,i}^\mathrm{T} \mathbf{w}_{k'})}
    \right)
\end{equation}

\noindent \textbf{Gradient \textit{w.r.t} classifier weights.} To analyze the learning behavior of the classifier, we first compute the gradient of $\mathcal{L}_{\rm CE}$ \textit{w.r.t} the classifier weights:
\setlength\abovedisplayskip{4pt}
\setlength\belowdisplayskip{4pt}
\begin{equation}
\label{sup:eq:2}
    \frac{\partial \mathcal{L}_{\rm CE}}{\partial \mathbf{w}_k} =
    \underbrace{\sum_{i=1}^{n_k} -(1-p_k(\mathbf{z}_{k,i})) \mathbf{z}_{k,i}}_{\rm pulling \ part} + 
    \underbrace{\sum_{k' \ne k}^K \sum_{j=1}^{n_{k'}} p_{k}(\mathbf{z}_{k',j}) \mathbf{z}_{k',j}}_{\rm forcing \ part}
\end{equation}
where $n_{k}$ denotes the number of training samples in the $k$-th class, and $p_k(\mathbf{z})$ is the predicted probability of $\mathbf{z}$ belongs to the $k$-th class, which is calculated with the softmax function:
\begin{equation}
    p_k(\mathbf{z}) = \frac{\exp (\mathbf{z}^{\mathrm{T}}\mathbf{w}_k)}{\sum_{k'=1}^K \exp (\mathbf{z}^{\mathrm{T}}\mathbf{w}_{k'})}, 1 \le k \le K
\end{equation}

In Eq. \ref{sup:eq:2}, we decompose the gradient \textit{w.r.t} the classifier weight $\mathbf{w}_k$ into two parts, \textit{the pulling part} and \textit{the forcing part}. The pulling part contains the effects of features from the same class (i.e., $\mathbf{z}_{k,i}$), which pulls the classifier weight towards the $k$-th feature cluster and each feature has an influence of $1-p_k(\mathbf{z}_{k,i})$. Meanwhile, the forcing part of the gradient contains the features from other classes to push $\mathbf{w}_k$ away from the wrong clusters, and each feature has an influence of $p_k(\mathbf{z}_{k',j})$. When the vanilla model is trained on biased datasets, the prevalent bias-aligned samples of the $k$-th class will dominate the pulling part of the gradient. The classifier weight $\mathbf{w}_k$ will be pulled towards the center of the bias-aligned features, which have a strong correlation between the class label $k$ and a bias attribute $b_k$. The biased feature space based on shortcut will thus be formed at the early period of training, as the result of the imbalanced magnitude of gradients across different attributes. Similarly, the forcing part of the gradient is also guided by the bias-aligned samples of other classes, further reinforcing the tendency of shortcut learning. It confirms our observation in Section \ref{sec:3.2} that the model's pursuit of shortcut correlation leads to a biased, non-collapsed feature space, which is hard to rectify in the subsequent training steps.

\noindent \textbf{Gradient \textit{w.r.t} features.} Furthermore, we compute the gradient of $\mathcal{L}_{\rm CE}$ \textit{w.r.t} the last-layer features:
\begin{equation}
\label{sup:eq:4}
    \frac{\partial \mathcal{L}_{\rm CE}}{\partial \mathbf{z}_{k,i}} =
    \underbrace{-(1-p_k(\mathbf{z}_{k,i})) \mathbf{w}_{k}}_{\rm pulling \ part} + 
    \underbrace{\sum_{k' \ne k}^K p_{k'}(\mathbf{z}_{k,i}) \mathbf{w}_{k'}}_{\rm forcing \ part}
\end{equation}

In Eq. \ref{sup:eq:4}, the gradient \textit{w.r.t} features is also considered as the combination of \textit{the pulling part} and \textit{the forcing part}. The pulling part represents the pulling effect of the classifier weight from the same class (i.e., $\mathbf{w}_k$), which will guide the features to align with the prediction behavior of the classifier. The forcing part, on the contrary, represents the pushing effect of other classifier weights. As we discussed before, the model's reliance on simple shortcuts is formed at the early stage of training, due to the misled classifier weights toward the centers of bias-aligned features. It results in a biased decision rule of the classifier, which directly affects the formation of the last-layer feature space. Consider the bias-conflicting samples of the $k$-th class, although the pulling part of the gradient supports its convergence towards the right classifier weight $\mathbf{w}_k$, the established shortcut correlation is hard to reverse and eliminate, which has been demonstrated with the metrics of Neural Collapse in Section \ref{sec:3.2}.

\subsection{Analysis of ETF-Debias}
In light of the analysis of vanilla training, our proposed debiasing framework, ETF-Debias, turns the easy-to-follow shortcut into the prime features, which guides the model to skip the active learning of shortcuts and directly focus on the intrinsic correlations. We have provided a brief theoretical justification of our method in Section \ref{sec:4.3}, and the detailed illustrations are as follows.

When trained on biased datasets, we assume each class $[1,...,K]$ is strongly correlated with a bias attribute $[b_1,...,b_K]$. 
Based on the mechanism of prime training, we denote the $i$-th feature of the $k$-th class as $\widetilde{\mathbf{z}}_{k,i}=[\mathbf{z}_{k,i}, \mathbf{m}_{i,b}] \in \mathbb{R}^{2 \times d}$, where $\mathbf{z}_{k,i} \in \mathbb{R}^{d}$ represents the learnable features, and $\mathbf{m}_{i,b} \in \mathbb{R}^{d}$ represents the prime features retrieved based on the bias attribute $b$ of the input sample. With the definition, a bias-aligned sample of the $k$-th class $\mathbf{x}_{k,i}$ will have its feature in form of $\widetilde{\mathbf{z}}_{k,i}=[\mathbf{z}_{k,i}, \mathbf{m}_{b_k}]$, and a bias-conflicting sample will have its feature as $\widetilde{\mathbf{z}}_{k,i}=[\mathbf{z}_{k,i}, \mathbf{m}_{b_{k'}}], k' \ne k$. 
To keep the same form, we denote the classifier weight as $\widetilde{\mathbf{w}}_{k}=[\mathbf{w}_{k}, \mathbf{a}_{k}] \in \mathbb{R}^{2 \times d}$, where $\mathbf{w}_{k} \in \mathbb{R}^{d}$ represents the weight for intrinsic correlations and $\mathbf{a}_{k} \in \mathbb{R}^{d}$ is the weight for shortcut features. In the detailed convergence result of Neural Collapse (Section \ref{sup:converge}), we observe that, due to the fixed prime features and their strong correlation with the bias attributes, $\mathbf{a}_{k}$ will quickly collapse into its bias-correlated prime feature $\mathbf{m}_{b_k}$, and can be viewed as constant after just a few steps of training. Thus the cross-entropy loss can be re-written as:
\setlength\abovedisplayskip{5pt}
\setlength\belowdisplayskip{5pt}
\begin{equation}
\label{sup:eq:5}
    \mathcal{L}_{\rm CE}(\widetilde{\mathbf{z}}_{k,i}, \widetilde{\mathbf{W}}) = -\log \left( \frac{\exp (\widetilde{\mathbf{z}}_{k,i}^\mathrm{T} \widetilde{\mathbf{w}}_k)}{\sum_{k'=1}^K \exp(\widetilde{\mathbf{z}}_{k,i}^\mathrm{T} \widetilde{\mathbf{w}}_{k'})}\right) 
    = -\log \left( \frac{\exp([\mathbf{z}_{k,i}, \mathbf{m}_{i,b}]^\mathrm{T}[\mathbf{w}_{k}, \mathbf{a}_k])}{\sum_{k'=1}^K [\mathbf{z}_{k,i}, \mathbf{m}_{i,b}]^\mathrm{T}[\mathbf{w}_{k'}, \mathbf{a}_{k'}]} \right)
\end{equation}

\noindent \textbf{Gradient \textit{w.r.t} classifier weights.} With the avoid-shortcut learning framework, we justify that the introduced prime mechanism implicitly plays the role of re-weighting, which weakens the mutual convergence between bias-aligned features and the classifier weights, and amplifies the learning of intrinsic correlations. We first analyze the gradient of $\mathcal{L}_{\rm CE}$ \textit{w.r.t} the classifier weights $\widetilde{\mathbf{W}}$:
\begin{align}
    \frac{\partial \mathcal{L}_{\rm CE}}{\partial \widetilde{\mathbf{w}}_k} 
    &= \sum_{i=1}^{n_k} -(1-p_k(\widetilde{\mathbf{z}}_{k,i}))\widetilde{\mathbf{z}}_{k,i} + \sum_{k' \ne k}^K \sum_{j=1}^{n_{k'}} p_k(\widetilde{\mathbf{z}}_{k',j})\widetilde{\mathbf{z}}_{k',j} \nonumber\\
    &= \sum_{i=1}^{n_k} -\left(1- \frac{\exp([\mathbf{z}_{k,i}, \mathbf{m}_{i,b}]^\mathrm{T}[\mathbf{w}_{k}, \mathbf{a}_k])}{\sum_{k'=1}^K \exp(\widetilde{\mathbf{z}}_{k,i}^\mathrm{T} \widetilde{\mathbf{w}}_{k'})} \right) \widetilde{\mathbf{z}}_{k,i} + \sum_{k' \ne k}^K \sum_{j=1}^{n_{k'}} \frac{\exp([\mathbf{z}_{k',j}, \mathbf{m}_{j,b'}]^\mathrm{T}[\mathbf{w}_{k}, \mathbf{a}_k])}{\sum_{k'=1}^K \exp(\widetilde{\mathbf{z}}_{k',j}^\mathrm{T} \widetilde{\mathbf{w}}_{k'})} \widetilde{\mathbf{z}}_{k',j}  \label{sup:eq:6} \\
    &\le \sum_{i=1}^{n_k}  -\left(1- \frac{\exp(\mathbf{z}_{k,i}^\mathrm{T}\mathbf{w}_{k})}{\sum_{k'=1}^K \exp(\widetilde{\mathbf{z}}_{k,i}^\mathrm{T}\widetilde{\mathbf{w}}_{k'})} - \frac{\exp(\mathbf{m}_{i,b}^\mathrm{T}\mathbf{a}_{k})}{\sum_{k'=1}^K \exp(\widetilde{\mathbf{z}}_{k,i}^\mathrm{T}\widetilde{\mathbf{w}}_{k'})} \right) \widetilde{\mathbf{z}}_{k,i} \nonumber\\
    &\ \ \ \ + \sum_{k' \ne k}^K \sum_{j=1}^{n_{k'}} \left( \frac{\exp(\mathbf{z}_{k',j}^\mathrm{T}\mathbf{w}_k)}{\sum_{k''=1}^K \exp(\widetilde{\mathbf{z}}_{k',j}^\mathrm{T} \widetilde{\mathbf{w}}_{k''})} + \frac{\exp(\mathbf{m}_{j,b'}^\mathrm{T}\mathbf{a}_k)}{\sum_{k''=1}^K \exp(\widetilde{\mathbf{z}}_{k',j}^\mathrm{T} \widetilde{\mathbf{w}}_{k''})} \right) \widetilde{\mathbf{z}}_{k',j} \label{sup:eq:7}\\ 
    &\le \underbrace{\sum_{i=1}^{n_k} - (1 - p_k^{(b)}(\mathbf{m}_{i,b}) - p_k^{(l)}(\mathbf{z}_{k,i}) ) \widetilde{\mathbf{z}}_{k,i}}_{\rm pulling \ part} + \underbrace{\sum_{k' \ne k}^K \sum_{j=1}^{n_{k'}} (p_k^{(b)}(\mathbf{m}_{j,b'}) + p_k^{(l)}(\mathbf{z}_{k',j}) )\widetilde{\mathbf{z}}_{k',j}}_{\rm forcing \ part} \label{sup:eq:8}
\end{align}
The predicted probabilities both satisfy $0< (1-p_k(\widetilde{\mathbf{z}}_{k,i}))<1$ and $0<p_k(\widetilde{\mathbf{z}}_{k',j})<1$, and the scaling step from Eq. \ref{sup:eq:6} to Eq. \ref{sup:eq:7} is based on the Jensen inequality. The terms $p_k^{(l)}$ and $p_k^{(b)}$ are respectively the predicted probabilities for class labels and bias attributes, calculated with softmax:
\begin{align}
\label{sup:eq:pk_1}
    p_k^{(l)}(\mathbf{z}_{k,i}) = \frac{\exp (\mathbf{z}_{k,i}^\mathrm{T}\mathbf{w}_{k})}{\sum_{k'=1}^K \exp ([\mathbf{z}_{k,i}, \mathbf{m}_{i,b}]^\mathrm{T}[\mathbf{w}_{k'}, \mathbf{a}_{k'}])}
\end{align}
\begin{equation}
\label{sup:eq:pk_2}
    p_k^{(b)}(\mathbf{m}_{i,b}) = \frac{\exp (\mathbf{m}_{i,b}^\mathrm{T}\mathbf{a}_{k})}{\sum_{k'=1}^K \exp ([\mathbf{z}_{k,i}, \mathbf{m}_{i,b}]^\mathrm{T}[\mathbf{w}_{k'}, \mathbf{a}_{k'}])}
\end{equation}

Based on the gradient \textit{w.r.t} classifier weights in Eq. \ref{sup:eq:8}, the probability $p_k^{(b)}\propto\exp(\mathbf{m}_{b}^\mathrm{T}\mathbf{a}_{k})$ re-weights the influence of different samples during optimization. Consider the features of the $k$-th class, a bias-aligned sample with its feature $\widetilde{\mathbf{z}}_{k,i}=[\mathbf{z}_{k,i}, \mathbf{m}_{b_k}]$ will have a much higher probability $p_k^{(b)}$, compared to a bias-conflicting sample of the same class with a feature as $\widetilde{\mathbf{z}}_{k,j}=[\mathbf{z}_{k,j}, \mathbf{m}_{b_{k'}}], k' \ne k$. Therefore, according to the influence of each feature in the pulling part of gradient $(1 - p_k^{(b)}(\mathbf{m}_{i,b}) - p_k^{(l)}(\mathbf{z}_{k,i}) )$, the bias-aligned samples will have their influence down-weighted, while the influence of bias-conflicting samples are up-weighted, which weakens the dominance of the prevalent bias-aligned samples in the pulling effect and prevents the formation of shortcut correlations. Meanwhile, according to the forcing part of the gradient \textit{w.r.t} $\mathbf{w}_k$, samples from other classes but have the bias-correlated attribute $b_k$ will have a relatively strong forcing effect on the weight $\mathbf{w}_k$, which also disturbs the learning of simple shortcuts and redirect the model's attention to the intrinsic, generalizable relations.

\noindent \textbf{Gradient \textit{w.r.t} features.} With the implicit re-weighting mechanism of gradients, the classifier weights are directed away from the simple shortcuts since the beginning of model training. We also compute the gradient of $\mathcal{L}_{\rm CE}$ \textit{w.r.t} the feature $\widetilde{\mathbf{z}}_{k,i}$:
\setlength\abovedisplayskip{3pt}
\setlength\belowdisplayskip{5pt}
\begin{align}
    \frac{\partial \mathcal{L}_{\rm CE}}{\partial \widetilde{\mathbf{z}}_{k,i}} 
    &= -(1- p_k(\widetilde{\mathbf{z}}_{k,i})) \mathbf{w}_k
 + \sum_{k' \ne k}^K p_{k'}(\widetilde{\mathbf{z}}_{k,i})\mathbf{w}_{k'} \nonumber\\
    &= -\left(1- \frac{\exp([\mathbf{z}_{k,i}, \mathbf{m}_{i,b}]^\mathrm{T}[\mathbf{w}_{k}, \mathbf{a}_k])}{\sum_{k'=1}^K \exp(\widetilde{\mathbf{z}}_{k,i}^\mathrm{T} \widetilde{\mathbf{w}}_{k'})} \right) \mathbf{w}_k + \sum_{k' \ne k}^K \frac{\exp([\mathbf{z}_{k,i}, \mathbf{m}_{i,b}]^\mathrm{T}[\mathbf{w}_{k'}, \mathbf{a}_{k'}])}{\sum_{k''=1}^K \exp(\widetilde{\mathbf{z}}_{k,i}^\mathrm{T} \widetilde{\mathbf{w}}_{k''})}\mathbf{w}_{k'} \label{sup:eq:11}\\
    &\le -\left(1- \frac{\exp(\mathbf{z}_{k,i}^\mathrm{T}\mathbf{w}_{k})}{\sum_{k'=1}^K \exp(\widetilde{\mathbf{z}}_{k,i}^\mathrm{T}\widetilde{\mathbf{w}}_{k'})} - \frac{\exp(\mathbf{m}_{i,b}^\mathrm{T}\mathbf{a}_{k})}{\sum_{k'=1}^K \exp(\widetilde{\mathbf{z}}_{k,i}^\mathrm{T}\widetilde{\mathbf{w}}_{k'})} \right) \mathbf{w}_k \nonumber\\
    &\ \ \ \ + \sum_{k' \ne k}^K \left( \frac{\exp(\mathbf{z}_{k,i}^\mathrm{T}\mathbf{w}_{k'})}{\sum_{k''=1}^K \exp(\widetilde{\mathbf{z}}_{k,i}^\mathrm{T} \widetilde{\mathbf{w}}_{k''})} + \frac{\exp(\mathbf{m}_{i,b}^\mathrm{T}\mathbf{a}_{k'})}{\sum_{k''=1}^K \exp(\widetilde{\mathbf{z}}_{k,i}^\mathrm{T} \widetilde{\mathbf{w}}_{k''})} \right) \mathbf{w}_{k'} \label{sup:eq:12}\\
    &\le \underbrace{-(1 - p_k^{(b)}(\mathbf{m}_{i,b}) - p_k^{(l)}(\mathbf{z}_{k,i})) \mathbf{w}_k}_{\rm pulling \ part}  + \underbrace{\sum_{k' \ne k}^K (p_{k'}^{(b)}(\mathbf{m}_{i,b}) + p_{k'}^{(l)}(\mathbf{z}_{k,i}))\mathbf{w}_{k'}}_{\rm forcing \ part} 
\label{eq:wrt_feat}
\end{align}
where the scaling step from Eq. \ref{sup:eq:11} to Eq. \ref{sup:eq:12} is based on the Jensen inequality. Similar to the analysis before, the predicted probability for bias attributes $p_k^{(b)}\propto\exp(\mathbf{m}_{b}^\mathrm{T}\mathbf{a}_{k})$ also performs re-weighting on the gradient \textit{w.r.t} features. With the prime feature $\mathbf{m}_{b_k}$, a bias-aligned sample of the $k$-th class will have a down-weighted influence in the pulling part towards the classifier weight $\mathbf{w}_k$, which avoids the dominance of bias-aligned features around the weight centers. In contrast, a bias-conflicting feature will be emphasized and have a magnified pulling effect towards the weight $\mathbf{w}_k$, which is crucial for the learning of intrinsic correlations. As to the forcing part of the gradient, a bias-conflicting sample with prime feature $\mathbf{m}_{b_{k'}}$ will obtain a strong pushing effect from the weight $\mathbf{w}_{k'}$, which forces it away from the wrong weight center and mitigate the spurious correlations. The implicitly adjusted pulling and forcing effects both contribute to the feature's convergence according to the intrinsic correlations. 

To sum up, through the theoretical justification from the perspective of gradients, the introduced prime mechanism implicitly re-weights the pulling and forcing effect of gradients. The down-weighted influence of bias-aligned samples weakens the trend of pursuing simple shortcuts, and the up-weighted influence of bias-conflicting samples enhanced the focus on intrinsic correlations, thus leading to an unbiased feature space with improved generalization capability. 

\section{Dataset Description and Implementation Details}
\label{sup:data}

\subsection{Datasets}

\noindent We use two synthetic datasets (i.e., Colored MNIST and Corrupted CIFAR-10) as well as three real-world datasets (i.e., Biased FFHQ, Dogs \& Cats, and BAR) to evaluate the debiasing performance of our proposed method. The illustrative examples of the datasets are displayed in Fig. \ref{sup:fig:syn}, \ref{sup:fig:bffhq}, \ref{sup:fig:bar}.

\noindent \textbf{Colored MNIST \cite{kim2019learning}. } As a modified version of the MNIST dataset \cite{mnist}, Colored MNIST introduces the color of digits as the bias attribute. The ten classes of digits are each correlated with a certain color (e.g., red for digit 0). We conduct experiments with the dataset used in \cite{lee2023revisiting, nam2020learning, lee2021learning, hwang2022selecmix} and set the ratio of bias-conflicting training samples as $\{0.5 \%, 1 \%, 2 \%, 5 \%\}$. The unbiased test set contains an equal number of bias-aligned and bias-conflicting samples.

\noindent \textbf{Corrupted CIFAR-10 \cite{hendrycks2018benchmarking}.} Also modified from the CIFAR-10 dataset \cite{krizhevsky2009learning}, Corrupted CIFAR-10 apply different types of corruptions to the original images, with each class correlated with a certain type of corruption. Following the previous studies \cite{tartaglione2021end,hwang2022selecmix}, we adopt the corruptions of \textit{Brightness, Contrast, Gaussian Noise, Frost, Elastic Transform, Gaussian Blur, Defocus Blur, Impulse Noise, Saturate} and set the ratio of bias-conflicting training samples as $\{0.5 \%, 1 \%, 2 \%, 5 \%\}$. The unbiased test set contains an equal number of bias-aligned and bias-conflicting samples.

\noindent \textbf{Biased FFHQ \cite{lee2021learning}.} Based on the Flickr-Faces-HQ dataset \cite{karras2019style} of face images, Biased FFHQ (BFFHQ) is constructed with the target attribute of age and the bias attribute of gender. The bias-aligned samples contain images of young females (i.e., ages ranging from 10 to 29) and old males (i.e., ages ranging form 40 to 59), and the bias-conflicting samples contain old females and young males. We set the ratio of bias-conflicting training samples as $\{0.5 \%, 1 \%, 2 \%, 5 \%\}$. The unbiased test set contains only bias-conflicting samples.

\noindent \textbf{Dogs \& Cats \cite{kim2019learning}.} First used in \cite{kim2019learning}, the Dogs \& Cats dataset is constructed with the target attribute of animal and the bias attribute of color (i.e., bright or dark color). The bias-aligned samples contain images of dark cats and bright dogs, while the bias-conflicting samples contain bright cats and dark dogs. We set the ratio of bias-conflicting training samples as $\{1 \%, 5 \%\}$, and the unbiased test set contains only bias-conflicting samples.

\noindent \textbf{BAR \cite{nam2020learning}.} The Biased Action Recognition (BAR) dataset contains six classes of actions (i.e., \textit{Climbing, Diving, Fishing, Vaulting, Racing, Throwing}), each correlated with a bias attribute of place (i.e., \textit{Rockwall, Underwater, WaterSurface, APavedTrack, PlayingField, Sky}). The bias-conflicting samples contain images with rare action-place pairs. We set the ratio of bias-conflicting training samples as $\{1 \%, 5 \%\}$, and the unbiased test set contains only bias-conflicting samples.

\renewcommand\thefigure{B.\arabic{figure}}
\setcounter{figure}{0}
\begin{figure*}[htbp]
    \subcaptionbox{Colored MNIST}{
    \centering
    \includegraphics[width=0.5\textwidth]{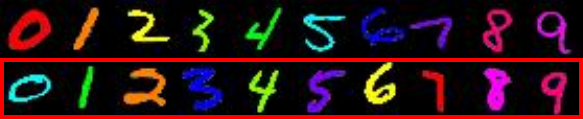}}
    \subcaptionbox{Corrupted CIFAR-10}{
    \centering
     \includegraphics[width=0.5\textwidth]{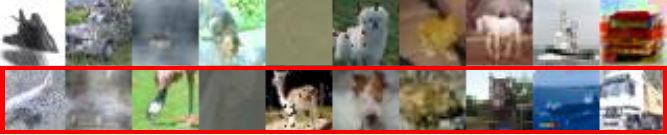}}
    \caption{Illustrative examples of synthetic datasets, (a) Colored MNIST with the bias attribute of color and (b) Corrupted CIFAR-10 with the bias attribute of different types of corruption (e.g. augmentations like Gaussian Blur and Pixelate, etc). Each column indicates a class in the dataset, the first row shows bias-aligned samples and the second row with red border shows bias-conflicting samples.}
    \label{sup:fig:syn}
\end{figure*}

\begin{figure}[htbp]
	\centering
	\subfloat[BFFHQ: \textit{young}]{\includegraphics[width=.16\columnwidth]{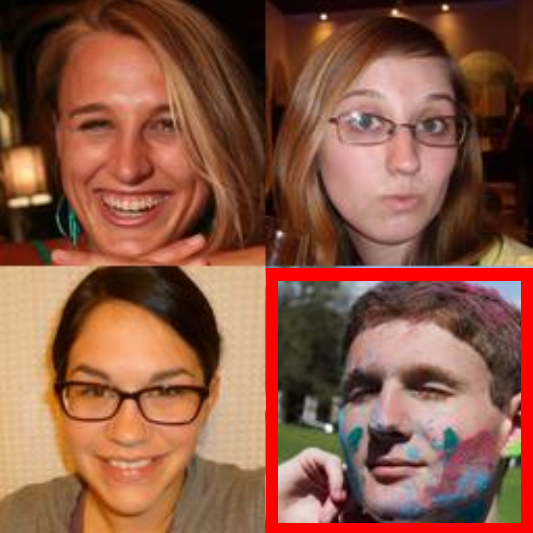}}\hspace{5pt}
	\subfloat[BFFHQ: \textit{old}]{\includegraphics[width=.16\columnwidth]{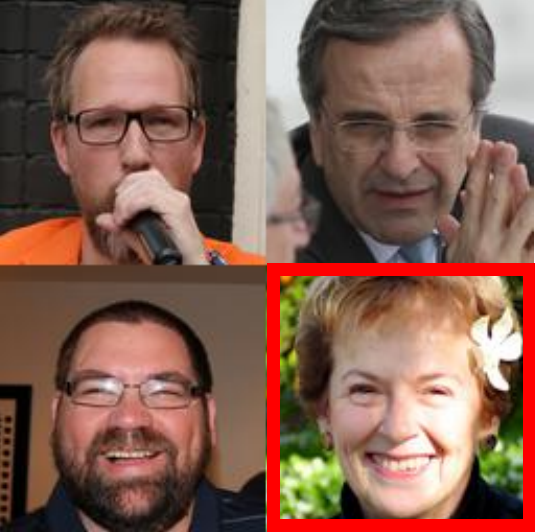}}\hspace{10pt}
 \subfloat[Dogs \& Cats: \textit{cat}]{\includegraphics[width=.16\columnwidth]{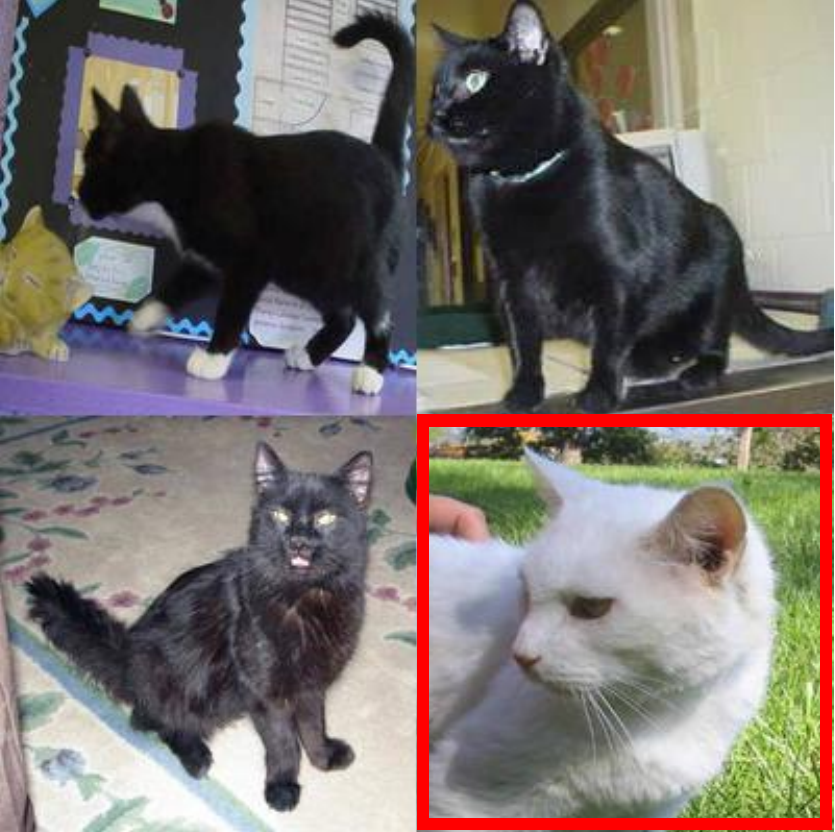}}\hspace{5pt}
	\subfloat[Dogs \& Cats: \textit{dog}]{\includegraphics[width=.16\columnwidth]{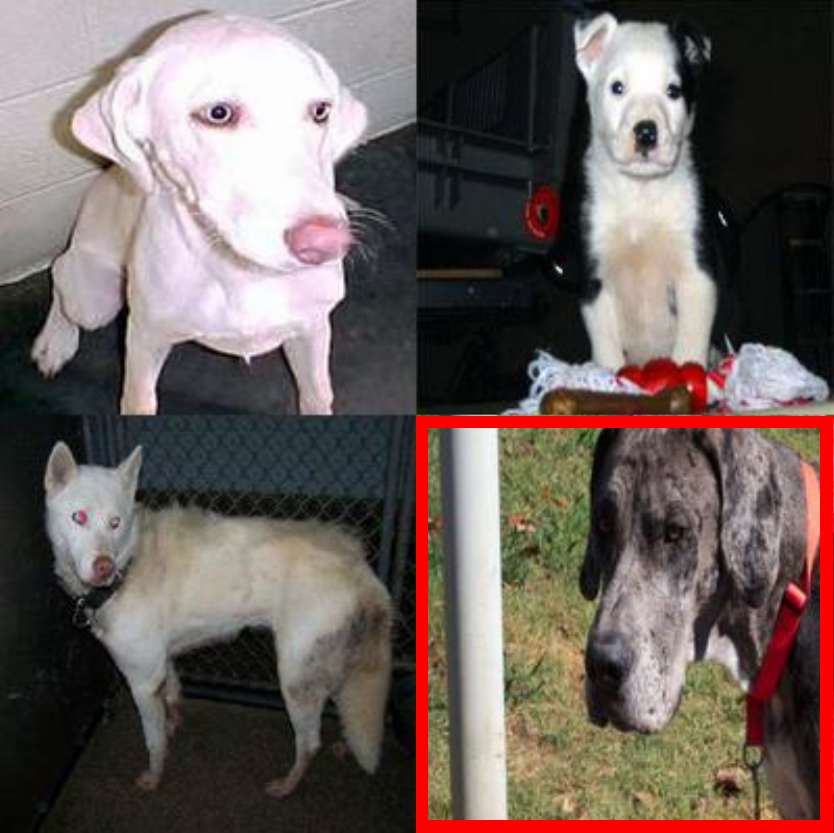}}\\
	\caption{Illustrative examples of (a)-(b) BFFHQ with the bias attribute of gender and (c)-(d) Dogs \& Cats with the bias attribute of color. (a) and (b) respectively indicate the class of young and old in BFFHQ, while (c) and (d) respectively indicate the class of cat and dog in Dogs \& Cats. The first three images in each sub-figure show bias-aligned samples and the last one with red border shows bias-conflicting samples.}
    \label{sup:fig:bffhq}
\end{figure}

\begin{figure}[htbp]
	\centering
	\subfloat[\textit{Climbing}]{\includegraphics[width=.15\columnwidth]{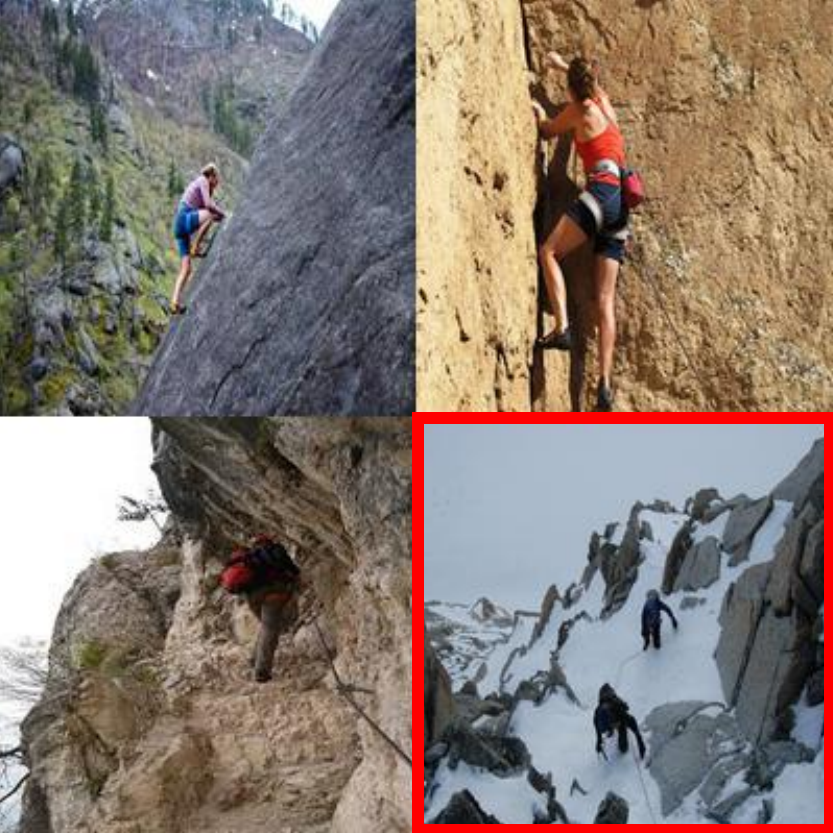}}\hspace{5pt}
	\subfloat[\textit{Diving}]{\includegraphics[width=.15\columnwidth]{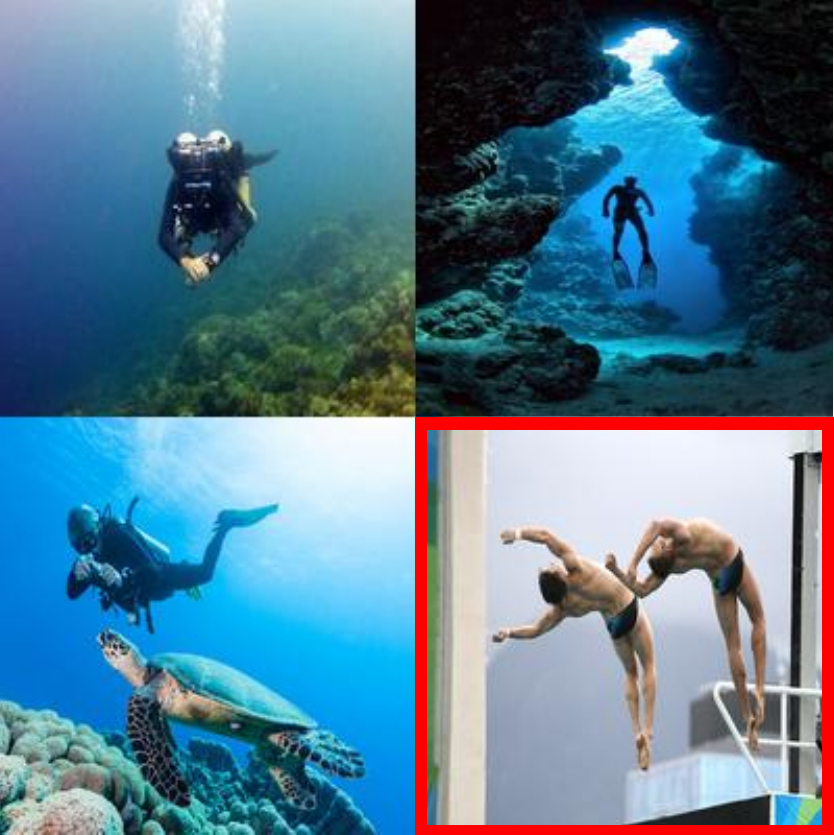}}\hspace{5pt}
 \subfloat[\textit{Fishing}]{\includegraphics[width=.15\columnwidth]{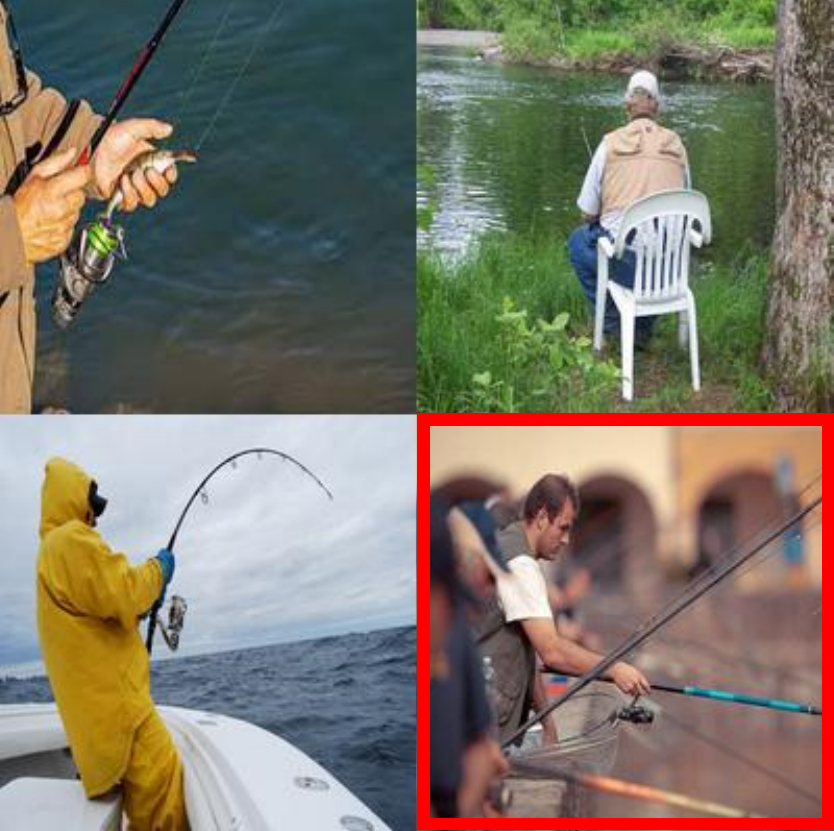}}\hspace{5pt}
  \subfloat[\textit{Racing}]{\includegraphics[width=.15\columnwidth]{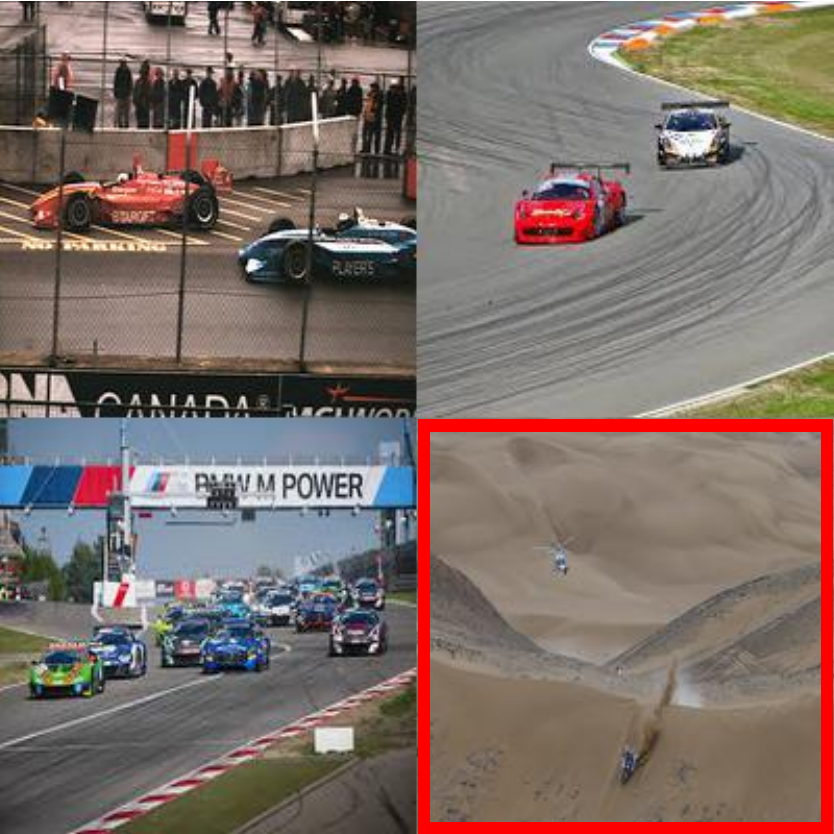}}\hspace{5pt}
   \subfloat[\textit{Throwing}]{\includegraphics[width=.15\columnwidth]{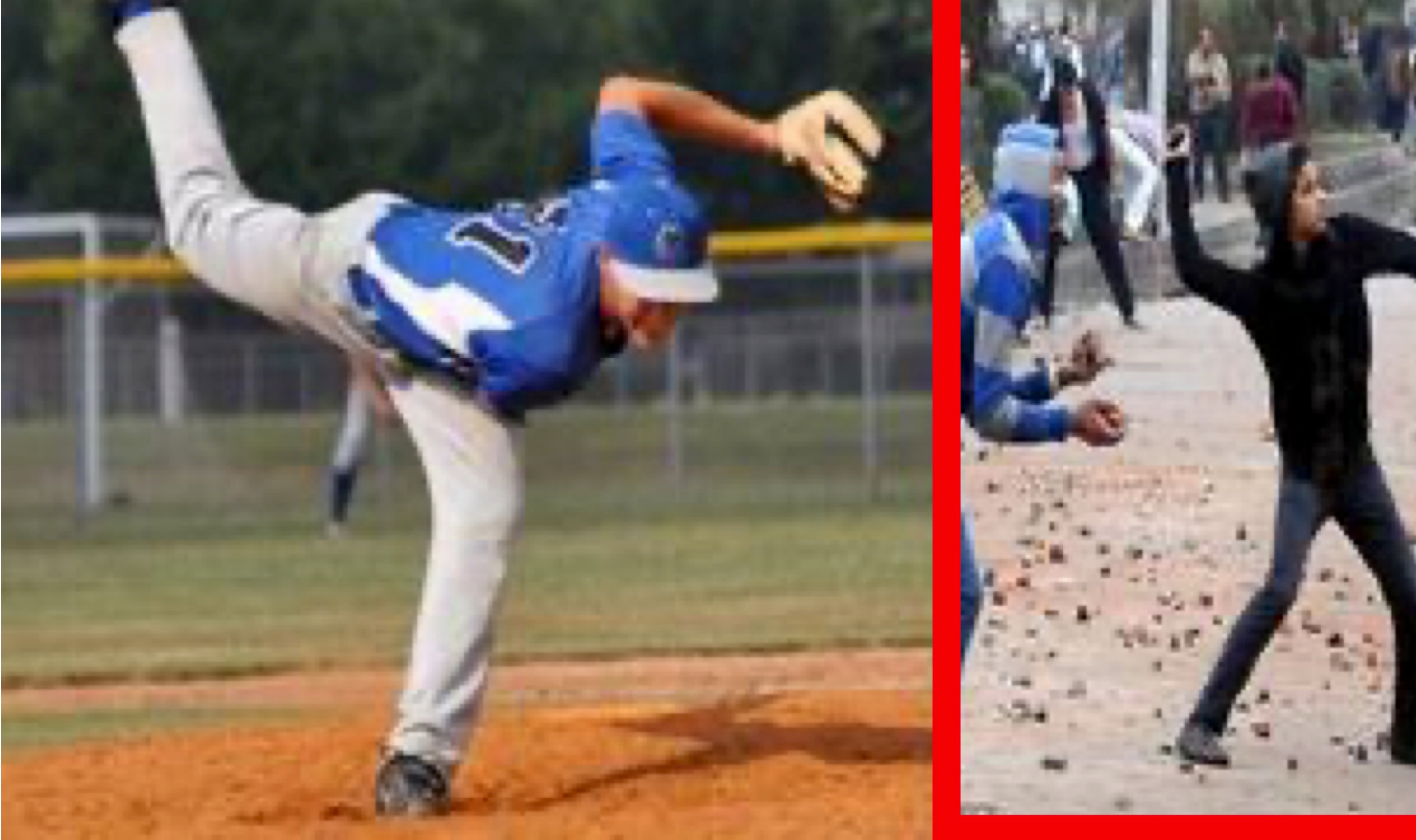}}\hspace{5pt}
	\subfloat[\textit{Vaulting}]{\includegraphics[width=.15\columnwidth]{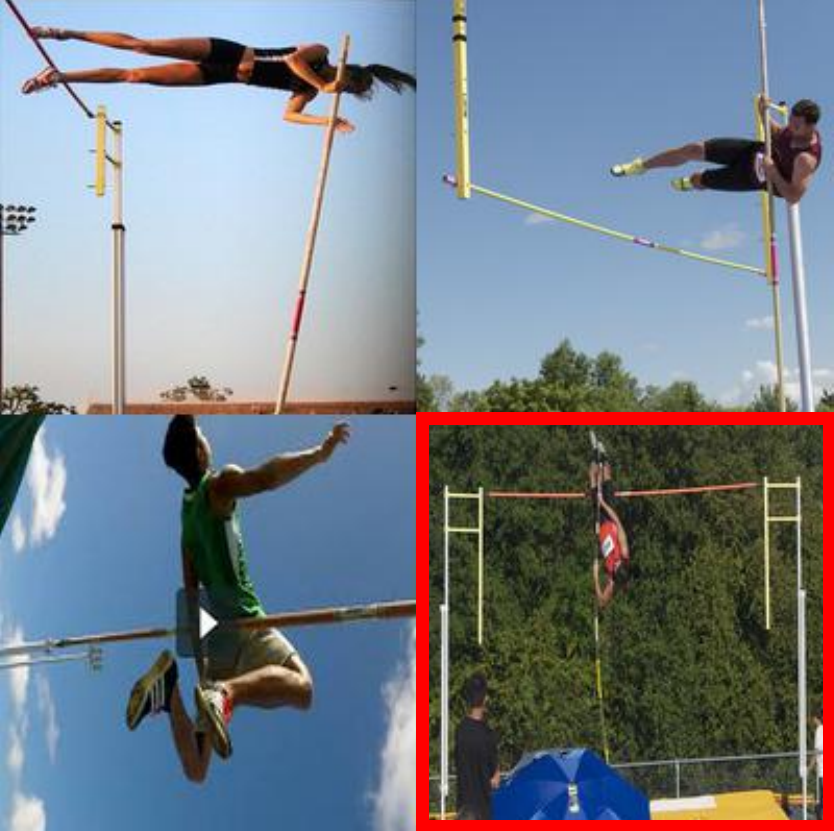}}\\
	\caption{Illustrative examples of BAR with the bias attribute of different places. (a)-(f) respectively indicates the classes of six actions. The first three images in each sub-figure show bias-aligned samples and the last one with red border shows bias-conflicting samples.}
    \label{sup:fig:bar}
\end{figure}

\subsection{Implementation Details}
During training, we set the batch size of 256 for Colored MNIST and Corrupted CIFAR-10, and 64 for BFFHQ, Dogs \& Cats and BAR. For all the baseline methods, we use the configuration in their official repositories. For ETF-Debias, we set the learning rate of 1e-3 and weight decay of 1e-5 for Colored MNIST, the learning rate of 1e-2 and weight decay of 2e-4 for Corrupted CIFAR-10, and the learning rate of 1e-4 and weight decay of 1e-6 for all the real-world datasets. The dimensions of learnable features and prime features are set as 100 and 64 for MLP and ResNet-20 respectively. The hyper-parameter $\alpha$ is set as 0.8, and all the evaluated models are trained for 200 epochs.
\section{More Convergence Results of Neural Collapse}
\label{sup:converge}

In Fig. \ref{fig:NC}, we display the trajectory of NC metrics when trained with ETF-Debias on the Corrupted CIFAR-10 dataset. Eliminating the obstacle of misled shortcut features, the debiased model exhibits better convergence properties during training, which forms a more symmetric feature space as on balanced datasets. We provide more convergence results of NC metrics on the synthetic biased dataset (Colored MNIST) and the real-world biased dataset (Dogs \& Cats) in Fig. \ref{sup:fig:NC_CMNIST} and Fig. \ref{sup:fig:NC_dog}. By applying ETF-Debias on various datasets with different types of bias attributes (\textcolor[RGB]{59,82,152}{\textbf{blue lines}}), we observe not only the better generalization capability on test sets, but also the more collapsed and robust structure of feature spaces (as indicated by the NC metrics), which provides empirical support for the effective guidance towards the intrinsic correlations. 
\renewcommand\thefigure{C.\arabic{figure}}
\setcounter{figure}{0}
\begin{figure*}[htbp]
\centering
\includegraphics[width=\columnwidth]{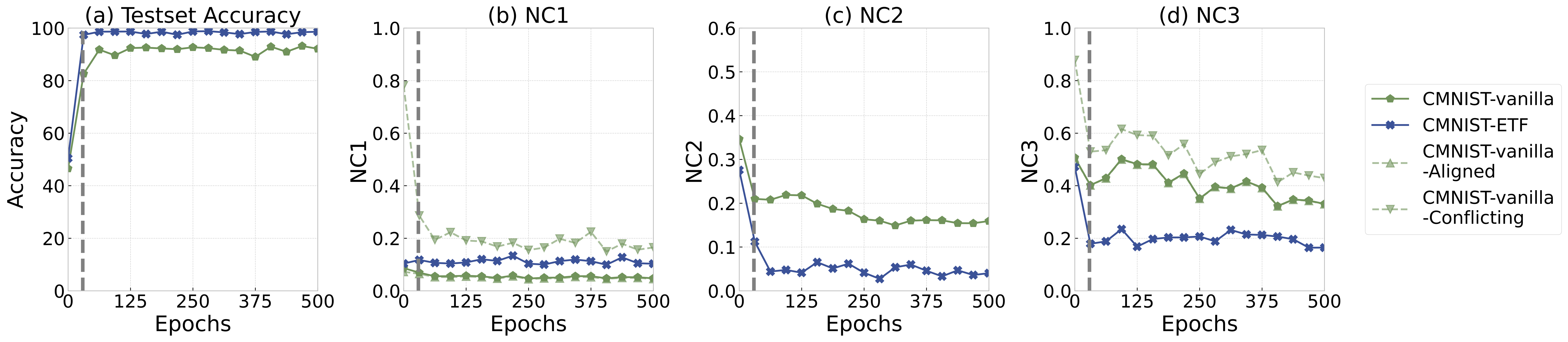}
\caption{Comparison of (a) testset accuracy and (b-d) Neural Collapse metrics on Colored MNIST with the bias ratio of 2.0\%. All vanilla models are trained with the standard cross-entropy loss for 500 epochs. The postfix \textit{-Aligned} and \textit{-Conflicting} indicate the results of bias-aligned and bias-conflicting samples respectively. The vertical dashed line at the epoch of 60 divides the two stages of training. All models are trained on ResNet-20. }
\label{sup:fig:NC_CMNIST}
\end{figure*}

\begin{figure*}[htbp]
\centering
\includegraphics[width=\columnwidth]{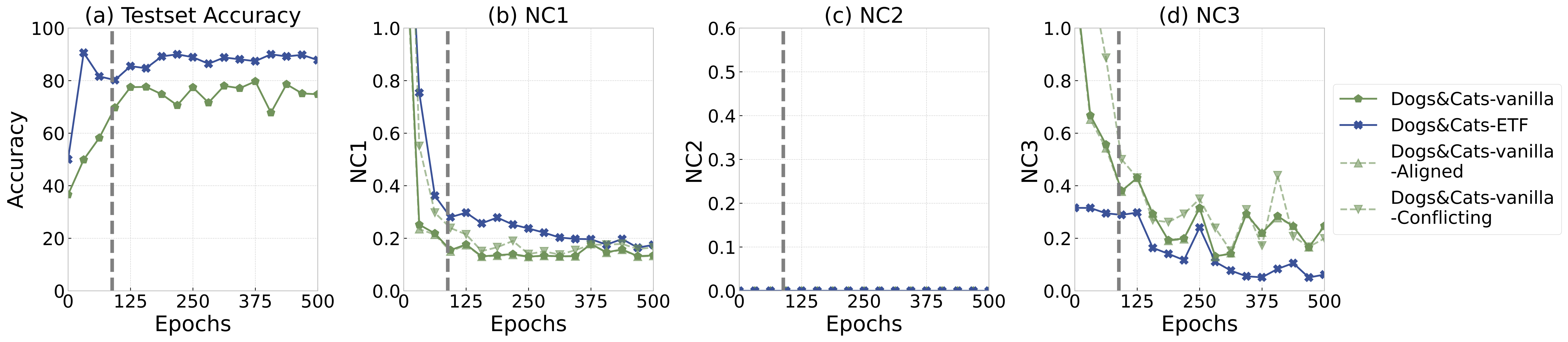}
\caption{Comparison of (a) testset accuracy and (b-d) Neural Collapse metrics on Dogs \& Cats with the bias ratio of 5.0\%. All vanilla models are trained with the standard cross-entropy loss for 500 epochs. The postfix \textit{-Aligned} and \textit{-Conflicting} indicate the results of bias-aligned and bias-conflicting samples respectively. The vertical dashed line at the epoch of 90 divides the two stages of training. Note that the Dogs \& Cats dataset is designed for the binary classification task, which means any classifier weights will follow the ETF structure (i.e., NC2 $\rightarrow$ 0), and the detailed theoretical support is provided in \cite{thrampoulidis2022imbalance}.}
\label{sup:fig:NC_dog}
\end{figure*}
\section{Detailed Results of BAR dataset}
\label{sup:bar}
Following the previous study \cite{nam2020learning}, we further report the class-wise accuracy on the unbiased test set of BAR. The debiasing performance on BAR with the bias ratio of 1.0\% and 5.0\% are displayed in Tab. \ref{sup:tab:bar1} and Tab. \ref{sup:tab:bar2} respectively. The compared baselines are the same as before.

\renewcommand\thetable{D.\arabic{table}}
\setcounter{table}{0}
\newcommand{\stdv}[1]{\scriptsize$\pm$#1}
\newcommand{\improv}[1]{\footnotesize\textcolor[RGB]{73,84,138}{\textbf{(+#1)}}}

\begin{table}[htbp]
\centering
\caption{Class-wise accuracy on the unbiased test set of BAR. The ratio of bias-conflicting training sample is 1.0\%. Best performances are marked in bold, and the number in brackets indicates the improvement compared to the best result in baselines.}
\label{sup:tab:bar1}
\begin{adjustbox}{width=0.95\textwidth}
\begin{tabular}{lcccccccl} 
\toprule[1.5pt]
Action & Climbing & Diving & Fishing & Vaulting & Racing & Throwing & & Average
\\
\cmidrule{1-7} \cmidrule{9-9}
Vanilla
& 75.46\stdv{1.68}
& 52.59\stdv{4.51} 
& 75.00\stdv{2.78} 
& 71.15\stdv{3.40} 
& 85.47\stdv{0.85} 
& 48.36\stdv{0.81} 
& & 68.00\stdv{0.43}

\\

LfF \cite{nam2020learning}
& {72.52\stdv{8.73}}
& {67.16\stdv{8.96}}
& {69.44\stdv{5.56}}
& {71.14\stdv{3.18}}
& {79.77\stdv{1.98}}
& {40.84\stdv{5.08}}
& & {68.30\stdv{0.97}}

\\

LfF+BE \cite{lee2023revisiting}
& {82.78\stdv{3.86}}
& {62.96\stdv{5.88}}
& {62.96\stdv{4.24}}
& {82.07\stdv{6.79}}
& {82.62\stdv{3.85}}
& {43.42\stdv{2.27}}
& & {71.70\stdv{1.33}}

\\

EnD \cite{tartaglione2021end}
& {79.85\stdv{2.29}}
& {50.62\stdv{3.08}}
& {73.15\stdv{3.21}}
& {74.51\stdv{0.49}}
& {82.05\stdv{0.00}}
& {49.29\stdv{4.88}}
& & {68.25\stdv{0.19}}

\\

SD \cite{zhang2023benign}
& {79.49\stdv{2.29}}
& {58.12\stdv{1.96}}
& {71.29\stdv{3.21}}
& {80.67\stdv{2.22}}
& {82.05\stdv{0.86}}
& {34.74\stdv{2.15}}
& & {67.73\stdv{0.35}}

\\

DisEnt  \cite{lee2021learning}
& {79.85\stdv{4.57}}
& {54.56\stdv{0.85}}
& {75.00\stdv{5.56}}
& {73.67\stdv{4.15}}
& {87.17\stdv{2.57}}
& {44.13\stdv{2.15}}
& & {69.30\stdv{1.27}}

\\

Selecmix \cite{hwang2022selecmix}
& {69.60\stdv{1.68}}
& {59.26\stdv{9.46}}
& {79.63\stdv{4.24}}
& {66.95\stdv{5.47}}
& {87.18\stdv{2.96}}
& {56.34\stdv{5.64}}
& & {69.83\stdv{1.02}}

\\

\textbf{ETF-Debias}
& {83.15\stdv{1.27}}
& {63.70\stdv{4.86}}
& {76.85\stdv{4.25}}
& {74.79\stdv{5.11}}
& {87.46\stdv{1.98}}
& {50.23\stdv{4.53}}
& & {\textbf{72.79}\stdv{0.21} \improv{1.09}}

\\
\bottomrule[1.5pt]
\end{tabular}
\end{adjustbox}
\end{table}

\begin{table}[htbp]
\centering
\caption{Class-wise accuracy on the unbiased test set of BAR. The ratio of bias-conflicting training sample is 5.0\%. Best performances are marked in bold, and the number in brackets indicates the improvement compared to the best result in baselines.}
\label{sup:tab:bar2}
\begin{adjustbox}{width=0.95\textwidth}
\begin{tabular}{lcccccccl}
\toprule[1.5pt]
Action & Climbing & Diving & Fishing & Vaulting & Racing & Throwing & & Average \\
\cmidrule{1-7} \cmidrule{9-9}
Vanilla
& 86.44\stdv{0.64}
& 85.19\stdv{0.75} 
& 79.63\stdv{3.20} 
& 82.63\stdv{1.75} 
& 89.85\stdv{0.19} 
& 52.11\stdv{6.13} 
& & 79.34\stdv{0.19}

\\

LfF \cite{nam2020learning}
& {84.24\stdv{2.54}}
& {84.69\stdv{4.93}}
& {75.92\stdv{8.93}}
& {80.67\stdv{2.22}}
& {89.74\stdv{0.86}}
& {52.58\stdv{6.50}}
& & {80.25\stdv{1.27}}

\\

LfF+BE \cite{lee2023revisiting}
& {86.77\stdv{2.92}}
& {85.96\stdv{1.91}}
& {78.30\stdv{5.33}}
& {82.11\stdv{5.76}}
& {90.16\stdv{1.29}}
& {55.70\stdv{1.95}}
& & {82.00\stdv{1.24}}

\\

EnD \cite{tartaglione2021end}
& {85.71\stdv{1.10}}
& {86.42\stdv{1.13}}
& {84.26\stdv{1.61}}
& {78.15\stdv{0.84}}
& {92.59\stdv{1.98}}
& {46.01\stdv{0.81}}
& & {78.86\stdv{0.36}}

\\

SD \cite{zhang2023benign}
& {84.25\stdv{0.64}}
& {82.91\stdv{4.12}}
& {82.41\stdv{1.60}}
& {86.27\stdv{1.28}}
& {88.03\stdv{0.86}}
& {50.70\stdv{1.41}}
& & {79.10\stdv{0.42}}

\\

Selecmix \cite{hwang2022selecmix}
& {84.98\stdv{5.08}}
& {79.51\stdv{2.38}}
& {81.48\stdv{5.78}}
& {85.43\stdv{0.97}}
& {92.02\stdv{0.50}}
& {49.30\stdv{3.73}}
& & {78.79\stdv{0.52}}

\\

DisEnt  \cite{lee2021learning}
& {84.98\stdv{1.68}}
& {89.13\stdv{5.66}}
& {77.77\stdv{0.00}}
& {80.39\stdv{2.95}}
& {90.59\stdv{1.71}}
& {48.82\stdv{1.63}}
& & {81.19\stdv{0.70}}

\\

\textbf{ETF-Debias}
& {90.48\stdv{1.68}}
& {87.65\stdv{6.39}}
& {85.18\stdv{1.61}}
& {93.28\stdv{2.22}}
& {92.31\stdv{2.26}}
& {53.05\stdv{0.81}}
& & {\textbf{83.66}\stdv{0.21} \improv{1.66}}

\\
\bottomrule[1.5pt]
\end{tabular}
\end{adjustbox}
\end{table}

\section{More Visualization Results}
\label{sup:vis}

In Section \ref{sec:5.4}, we provide the comparison of visualization results between vanilla training and ETF-Debias on Corrupted CIFAR-10, Dogs \& Cats, and BFFHQ. To further demonstrate the rectified attention of debiased models, we supplement more visualization results in Fig. \ref{sup:fig:vis}(a)-(d).

By explicitly visualizing the model's attention regions, we observe that ETF-Debias does encourage the model to focus on the intrinsic correlations rather than shortcuts. On synthetic biased datasets like Colored MNIST and Corrupted CIFAR-10, the vanilla models are easily misled by artificial shortcuts and focus on the non-essential parts of images, as shown in Fig. \ref{sup:fig:vis}(a) \& (c). In constrast, when the shortcut learning is suppressed by prime features, the debiased model trained with ETF-Debias will pay attention to the digits and objects themselves and make unbiased classification. On real-world biased datasets with diverse bias attributes, the vanilla models tend to focus on the background or the distinctive part of the bias attributes. For example, the vanilla models rely on the distinctive part of gender such as the beard of a male or the accessories of a female in Fig. \ref{sup:fig:vis}(d). The pursuit of shortcut correlations has also been prevented with ETF-Debias, which guides the debiased model to focus more on facial features and predict the target attribute.

\renewcommand\thefigure{E.\arabic{figure}}
\setcounter{figure}{0}
\begin{figure*}[htbp]

\subfloat[Colored MNIST\\ \textit{bias:color}]{
    \centering
\includegraphics[width=0.5\columnwidth]{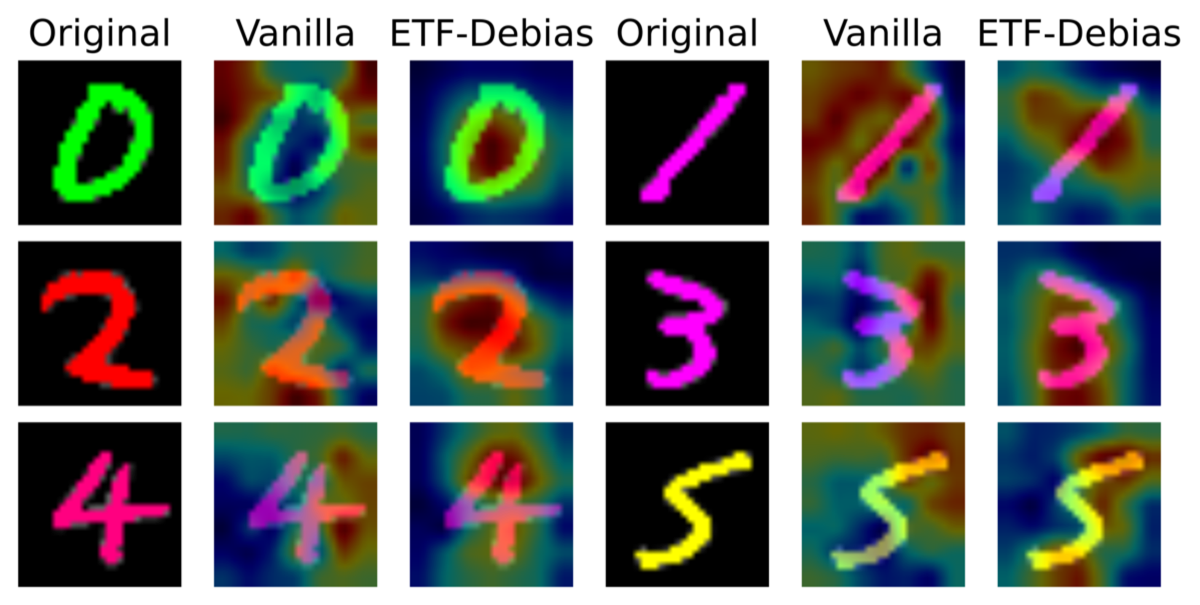}
}
\hfill 
\subfloat[BAR\\ \textit{bias:place}]{
    \centering
\includegraphics[width=0.5\columnwidth]{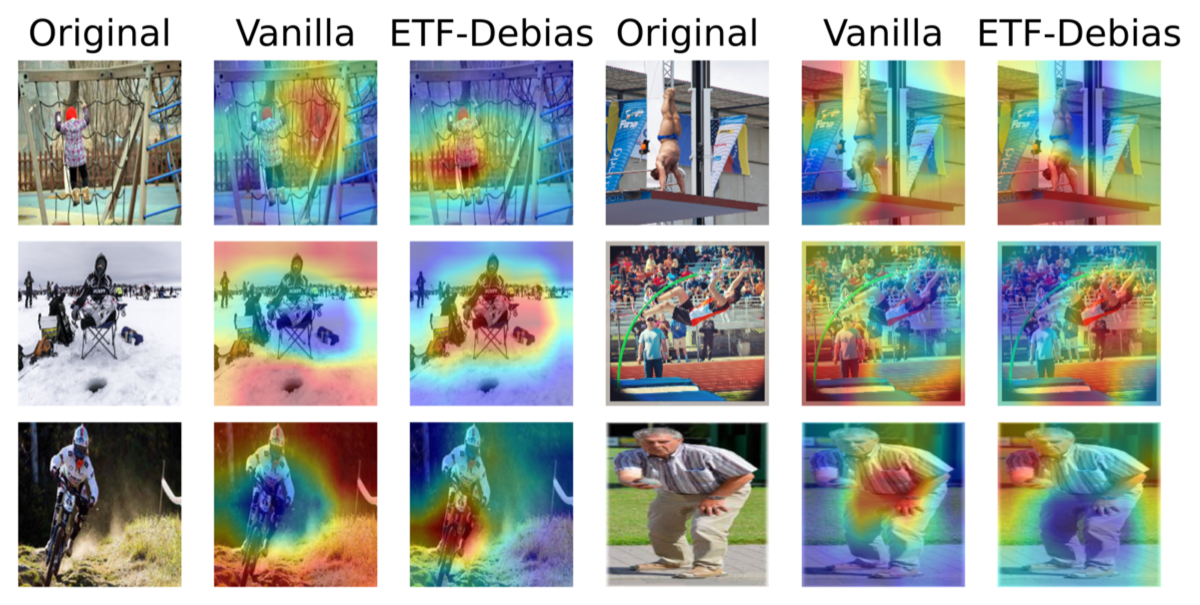}
}
\newline
\subfloat[Corrupted CIFAR-10\\ \textit{bias:corruption}]{
    \centering
\includegraphics[width=0.5\columnwidth]{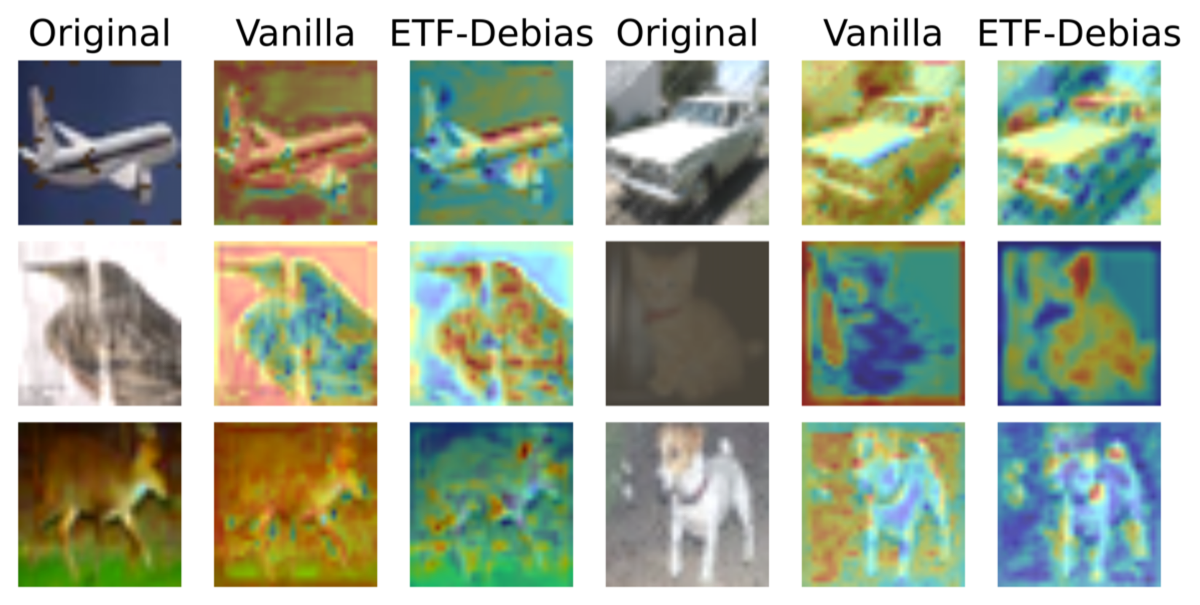}
}
\hfill 
\subfloat[Biased FFHQ\\ \textit{bias:gender}]{
    \centering
\includegraphics[width=0.5\columnwidth]{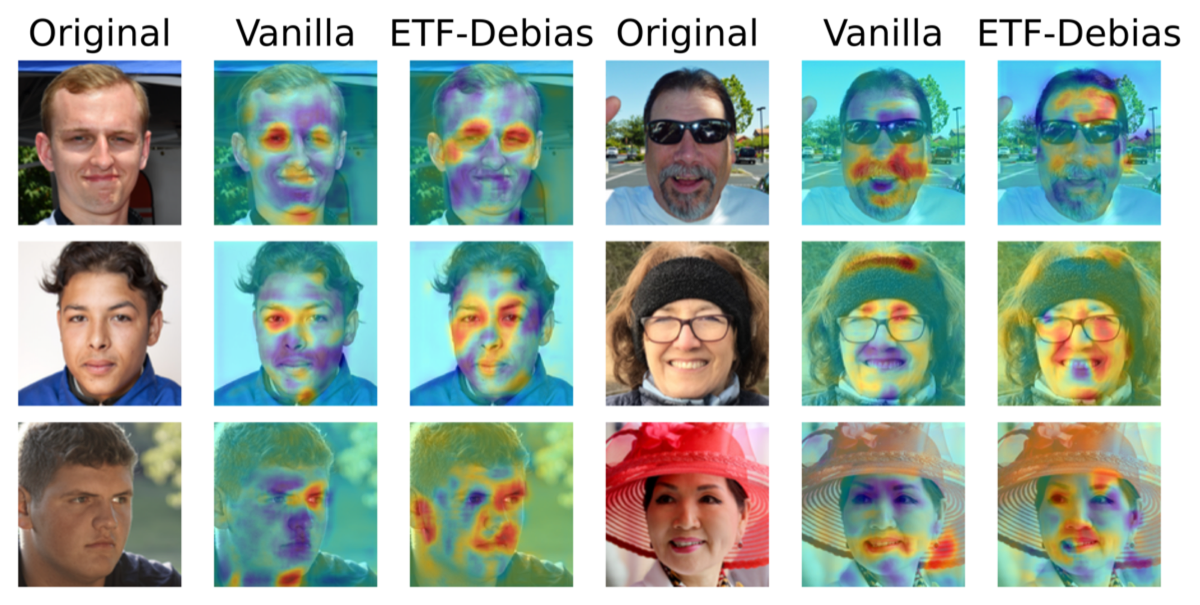}
}

    \caption{More comparison of visualization results on bias-conflicting samples. We display the result of CAM \cite{zhou2016learning} on (a) Colored MNIST, with the bias attribute as the color of digits, (b) BAR, with the bias attribute as the background of images, (c) Corrupted CIFAR-10, with the bias attribute as different types of corruptions on the entire image, (d) Biased FFHQ, with the bias attribute as gender. The original images in (a)-(c) represent the first six classes in each dataset. The original images in the first column of (d) are from the class of \textit{young} in BFFHQ, while the original images in the fourth column are from the class \textit{old}. All models are trained on ResNet-20 to obtain the CAM results.}
\label{sup:fig:vis}
\end{figure*}

\end{document}